\documentclass[10pt,twocolumn,letterpaper]{article}

\usepackage[pagenumbers]{cvpr}

\usepackage{amsfonts,amsthm,mathtools}

\usepackage[utf8]{inputenc}
\usepackage[T1]{fontenc}
\usepackage{nicefrac}
\usepackage{microtype}
\usepackage{multirow}
\usepackage{wrapfig}

\usepackage{listings}
\definecolor{lstbg}{HTML}{F7F7F9}
\definecolor{lstkw}{HTML}{AA3731}
\definecolor{lststr}{HTML}{448C27}
\definecolor{lstcom}{HTML}{7D8B99}
\lstdefinestyle{pythonratio}{
  language=Python,
  basicstyle=\ttfamily\footnotesize,
  keywordstyle=\color{lstkw}\bfseries,
  stringstyle=\color{lststr},
  commentstyle=\color{lstcom}\itshape,
  backgroundcolor=\color{lstbg},
  frame=single,
  framerule=0pt,
  rulecolor=\color{lstbg},
  xleftmargin=6pt,
  xrightmargin=6pt,
  aboveskip=6pt,
  belowskip=6pt,
  breaklines=true,
  showstringspaces=false,
  columns=fullflexible,
  keepspaces=true,
}
\usepackage{algorithm}
\usepackage{algorithmic}

\usepackage{mdframed}
\mdfdefinestyle{mainthm}{
    linewidth=0.5pt,
    linecolor=black!40,
    backgroundcolor=black!3,
    innerleftmargin=10pt,
    innerrightmargin=10pt,
    innertopmargin=4pt,
    innerbottommargin=4pt,
    skipabove=6pt,
    skipbelow=6pt,
}

\definecolor{cvprblue}{rgb}{0.21,0.49,0.74}
\usepackage[pagebackref,breaklinks,colorlinks,allcolors=cvprblue]{hyperref}
\usepackage[capitalize]{cleveref}

\theoremstyle{plain}
\newtheorem{theorem}{Theorem}
\newtheorem{proposition}{Proposition}

\newtheorem*{restatedresult}{\restatedname}
\newcommand{\restatedname}{}
\newenvironment{restated}[2]{\renewcommand{\restatedname}{#1}\begin{restatedresult}[#2]}{\end{restatedresult}}
\theoremstyle{definition}
\newtheorem{definition}{Definition}
\newtheorem{assumption}{Assumption}
\theoremstyle{remark}

\definecolor{takeawayBg}{HTML}{EEF6FF}
\definecolor{takeawayRule}{HTML}{2F6FD6}

\definecolor{designChoiceBg}{HTML}{FFF1F1}
\definecolor{designChoiceRule}{HTML}{C43B3B}

\newcommand{\EE}{\mathbb{E}}
\newcommand{\RR}{\mathbb{R}}

\definecolor{colpstar}{HTML}{017100}
\definecolor{colq}{HTML}{D64541}

\newcommand{\pstar}{p^{*}}

\newcommand{\q}{{q}}
\newcommand{\qstar}{q^{*}}

\newenvironment{widefigure}[1][t]{\begin{figure*}[#1]}{\end{figure*}}
\newenvironment{widetable}[1][t]{\begin{table*}[#1]}{\end{table*}}
\usepackage{adjustbox}
\usepackage{colortbl}
\usepackage{needspace}
\definecolor{mmoktint}{HTML}{EAF2FB}
\definecolor{stdgray}{HTML}{707070}
\newcommand{\std}[1]{{\scriptsize\color{stdgray}$\pm#1$}}
\newcolumntype{M}{>{\columncolor{mmoktint}}c}

\AtEndPreamble{
  \crefname{assumption}{Assumption}{Assumptions}
  \Crefname{assumption}{Assumption}{Assumptions}
  \crefname{proposition}{Prop.}{Props.}
  \crefname{theorem}{Thm.}{Thms.}
  \crefname{algorithm}{Alg.}{Algs.}
  \crefname{appendix}{App.}{Apps.}
  \crefname{subappendix}{App.}{Apps.}
  \crefname{subsubappendix}{App.}{Apps.}
}

\def\confName{CVPR}
\def\confYear{2027}

\title{Mirrored Min-of-K Training}

\author{Jiaru Zhang\thanks{Correspondence to: Jiaru Zhang. This research was supported by the NVIDIA Academic Grant Program using its
remote GPU server.}~, Zeyun Deng, Hyunsoo Yun, Yupeng Zhou, Bingyao Du, Juntong Peng, \\ Zichong Yang, Can Cui, Chuheng Wei, Ruqi Zhang, Ziran Wang\\
Purdue University\\
\tt{\small\{jiaru, ruqiz, ziran\}@purdue.edu}
}

\begin{document}
\maketitle

\begin{abstract}
Min-of-$K$ (MoK) training is widely used to train models that produce diverse outputs from random noise.
For each target, it draws $K$ outputs from the model and backpropagates only through the one closest to the target.
However, MoK generally does not make the generated distribution match the data distribution.
In this paper, we analyze the MoK bias across all values of $K$ and show that tuning $K$ shifts between failure modes, from mode dropping at small $K$ to distribution flattening at large $K$.
We further prove that every nontrivial loss based only on target--candidate distances fails to recover some data distribution, since these distances do not reveal how the candidates are placed relative to each other.
Based on these analyses, we propose \textbf{mirrored MoK (MMoK)}, a drop-in modification of MoK training that draws one extra sample from the model, applies the closest-candidate selection rule against it, and subtracts the resulting distance.
Since this distance is measured between samples of the model, it provides additional information about how the candidates are placed relative to each other.
We prove that MMoK can in principle recover data distributions that MoK cannot.
We empirically validate MMoK on five scenarios, namely synthetic mixtures, pedestrian and vehicle trajectory prediction, tropical cyclone forecasting, robotic policy learning, and ambiguous medical image segmentation.
The results confirm the effectiveness and generality of MMoK in both distribution learning and task performance.
\end{abstract}

\section{Introduction}
\label{sec:intro}

\begin{figure}[t]
\centering
\includegraphics[width=\linewidth]{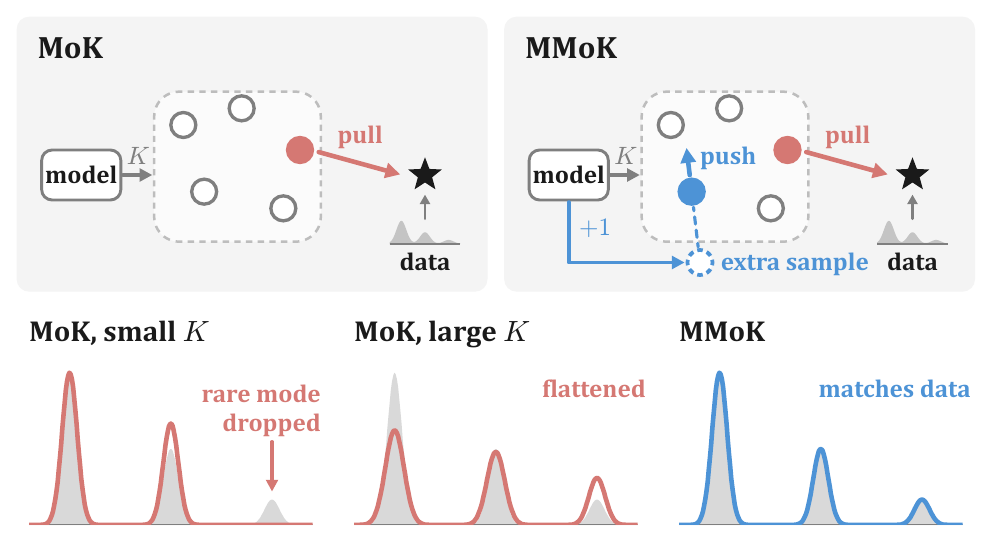}
\caption{\textbf{Top:} MoK pulls the candidate closest to the data point ($\bigstar$) toward it (red); MMoK additionally pushes the candidate closest to an extra model sample (dashed) away from it (blue).
\textbf{Bottom:} MoK drops the rare mode at small $K$ and flattens the distribution at large $K$, while MMoK matches the data (gray).}
\label{fig:motivation}
\end{figure}

Min-of-$K$ (MoK) training is a widely used paradigm for training models to produce diverse outputs~\citep{fan2017pointset,gupta2018social}.
As illustrated in \cref{fig:motivation} (top left), in each step, the model draws $K$ candidate outputs independently, and only the candidate closest to the target receives a gradient, which encourages the model to place its generated samples diversely.
Moreover, MoK is simple to implement because it only requires forwarding the model multiple times and backpropagating through the best one.
It has been applied to trajectory prediction~\citep{gupta2018social}, point-set generation~\citep{fan2017pointset}, tropical cyclone forecasting~\citep{huang2025tropicyclonenet}, robotic policy learning~\citep{rana2025imle,gladstone2026xm}, and ambiguous medical image segmentation~\citep{rakic2025tychestochasticincontextlearning}.

However, MoK training generally cannot make the model's samples follow the data distribution.
With a small $K$, the generated distribution that minimizes the MoK loss concentrates on the more frequent modes of the data distribution and drops rare ones~\citep{bennett2026freedom}.
With a large $K$, this distribution is flattened.
For instance, in one dimension and with the unsquared distance, its density becomes proportional to the square root of the data density~\citep{thiede2019variety}.
To correct this bias, \citet{thiede2019variety} proposed a post-hoc correction that raises a density estimate of the model's samples to a power tuned on held-out data.
However, this correction is limited in practice.
It requires many model samples and a density estimate for every input at inference, which introduces extra costs and difficulty in real-time settings.
In addition, it can only reshape the density the model already produces, so a mode that the model has dropped can never be recovered.

In this paper, we first analyze the MoK bias across all values of $K$, building on existing results~\citep{shahroudi2024evaluation,bennett2026freedom,zador1982asymptotic,thiede2019variety}.
As illustrated in \cref{fig:motivation} (bottom), no choice of $K$ guarantees recovery of the data distribution.
We further prove that every nontrivial loss based only on target--candidate distances fails to recover some data distribution.
Such a loss uses only distances to the current target and does not reveal how the candidates are placed relative to each other.
Therefore, for a target in one mode, the loss cannot tell whether the remaining candidates cover the other modes.
This calls for a loss that looks beyond the current target and directly probes the generated distribution.

Based on the analyses above, we propose \textbf{mirrored MoK (MMoK)}, a drop-in modification of the MoK training objective.
As illustrated in \cref{fig:motivation} (top right), in each training step, it draws one extra sample from the model, with no gradient flowing through it, and uses this sample as a second target.
The same $K$ candidates are matched to this extra sample using the same closest-candidate selection rule, and the resulting distance is subtracted from the original loss.
Intuitively, the MoK term pulls a candidate toward the data point, and the subtracted term pushes a candidate away from the extra sample.
The subtracted term mirrors the MoK loss, with the extra sample in place of the data point.
Since the subtracted distance is measured between samples of the model, it reveals how the candidates are placed relative to each other, which a target-only loss cannot.
When the generated distribution equals the data distribution, the two distance terms have the same expected gradient and cancel.
In practice, MMoK adds only one extra sample per step and does not require new hyperparameters or additional networks.
Unlike the post-hoc correction, MMoK changes only the training loss, so it does not need a density estimate or extra samples at inference, and it corrects the trained model itself.

We prove that a model whose generated distribution equals the data distribution is a stationary point of MMoK training.
In a simplified model of training on equidistant point modes, we further prove that this is the only stationary point, which shows that MMoK can recover data distributions that MoK cannot.
We conduct experiments on five scenarios, namely synthetic mixtures, pedestrian and vehicle trajectory prediction, tropical cyclone forecasting, robotic policy learning, and ambiguous medical image segmentation.
MMoK generally outperforms standard MoK, as well as the post-hoc correction in most settings.
For tropical cyclone forecasting, MMoK reduces the probabilistic track error from $144.5$\,km to $97.0$\,km and widens the spread of pressure forecasts from $0.20$ to $0.97$ times their error.
On ETH/UCY pedestrian trajectories, MMoK assigns the highest likelihood to the true future in three of the five scenes and is more accurate than the post-hoc correction in all five.
Both advantages also hold on NGSIM highway trajectories at $K = 8$ and $K = 20$.
For QUBIQ brain-growth segmentation at $K = 8$, MMoK's predicted masks vary about as much as the rater masks, while MoK's vary only about half as much.

In summary, our contributions are as follows:
\begin{itemize}
    \item We analyze the MoK bias across all values of $K$ and prove that every nontrivial target-only loss fails to recover some data distribution.
    \item We propose MMoK, a drop-in modification, and prove stationarity at the data distribution for any $K$ and any distance and its uniqueness in a simplified model.
    \item We evaluate MMoK on five scenarios and show that it generally outperforms standard MoK and, in most settings, the post-hoc correction.
\end{itemize}

\section{Related work}
\noindent\textbf{Min-of-$K$ training.}
MoK is a widely used training paradigm in deep learning that draws multiple outputs as independent samples from the model and backpropagates only through the one closest to the target.%
\footnote{Multiple choice learning~\citep{guzman2012mcl,lee2016smcl,rupprecht2017learning,letzelter2024winner} applies the same rule to a fixed set of predictors or output heads, a setting this paper does not consider.}
An early use is point-set generation, where it lets the model output a distribution that reflects the ambiguity of the target~\citep{fan2017pointset}.
It has since been used in many tasks, including trajectory prediction~\citep{gupta2018social,huang2020diversitygan}, human motion prediction~\citep{barquero2023belfusion},
tropical cyclone forecasting~\citep{huang2025tropicyclonenet}, diverse imitation learning~\citep{vahabpour2025sog}, robotic policy learning~\citep{rana2025imle,gladstone2026xm}, ambiguous medical image segmentation~\citep{rakic2025tychestochasticincontextlearning}, and vision-language-action policies~\citep{hosseinkhani2026imlevla}.
In these applications, MoK is usually regarded as a simple and effective way to produce diverse outputs and avoid mode collapse.

\noindent\textbf{Bias of min-of-$K$ and its correction.}
Several works have shown that MoK training is biased and does not generally make the generated distribution match the data distribution.
\citet{thiede2019variety} show that, in one dimension and at large $K$, the generated distribution that minimizes the MoK loss is flatter than the data distribution.
\citet{bennett2026freedom} analyzes discrete outputs at every $K$ and shows that this distribution concentrates on the more frequent outputs at small $K$ and becomes flatter as $K$ grows.
In the limit of infinite $K$, the MoK loss only requires the generated samples to cover the support of the data distribution and does not determine how the mass is distributed over it~\citep{gladstone2026xm,bennett2026freedom}.
This bias has been observed in practice~\citep{kothari2023safety}.
To remove the bias, \citet{thiede2019variety} also propose a post-hoc correction that reshapes the density at inference.
It needs to estimate a density from many samples for every input, which introduces additional computational cost at inference.
Moreover, the correction cannot restore modes that the model has dropped, as the trained model itself remains unchanged.

\noindent\textbf{Losses with a model sample as a second target.}
The idea of using a model sample as a second target has appeared in several contexts.
The energy score compares model samples both with the target and with each other and is strictly proper with the Euclidean distance~\citep{gneiting2007strictly,szekely2013energy}, so training with it targets the data distribution~\citep{bouchacourt2016disco,shen2025engression}.
Drifting models~\citep{deng2026drifting} attract generated samples toward data samples and repel them from other generated samples, so their update vanishes when the two distributions coincide.
With $K = 1$, MMoK has the same expected gradient as training with the energy score built from the same distance.
In this sense, MMoK is a natural extension of this idea to the MoK setting, which is widely used in practice.
\section{Analysis of the min-of-$K$ training}
\label{sec:map}
\subsection{Preliminaries}
\label{sec:preliminaries}

\noindent\textbf{Min-of-$K$ training.}
We consider MoK training of a model with target data distribution $\pstar$.
At each training step, the model produces $K$ candidate outputs $\hat y_{1:K} \sim \q$ i.i.d.\ from its generated distribution $\q$, compares them with one data point $x \sim \pstar$, and
 is trained with the loss on the best match alone:
\begin{equation}\label{eq:minofk}
  \mathcal{L}_{\mathrm{MoK}}
  \;=\;
  \min_{i \in \{1, \dots, K\}}\, \mathcal{D}(\hat y_i, x),
\end{equation}
where $\mathcal{D}$ is a distance between a candidate and the data point.
In our theoretical analysis, we mainly focus on $\mathcal{D}(u, v) = \|u - v\|^{r}$ with distance power $r > 0$, typically $r = 1$ or $r = 2$.
For conditional models, $\pstar$ and $\q$ are distributions conditioned on the same input, and every statement of the paper holds for each conditioning input.

Note that MoK uses the selection rule in \cref{eq:minofk} only during training.
At inference, the model simply draws samples from $\q$.
Therefore, it is natural to ask: \emph{does minimizing the training objective make $\q$ converge to $\pstar$?}
In general, it does not, and we show how the gap depends on $K$ below.

\subsection{Analysis across $K$}
\noindent\textbf{The optimal sampling distribution.}
To assist the analysis, we write the per-step loss \cref{eq:minofk} in expectation form and define $\qstar$ as the distribution that minimizes it over all $\q$:
\begin{equation}\label{eq:qstar}
  \qstar
  \;=\;
  \arg\min_{\q}\; \EE_{x \sim \pstar}\, \EE_{\hat y_{1:K} \sim \q}
  \Bigl[\, \min_{i \le K}\, \mathcal{D}(\hat y_i, x) \Bigr].
\end{equation}
We now analyze how $\qstar$ differs from $\pstar$ across $K$.

\noindent\textbf{Analysis in the $K = 1$ case.}
It is straightforward to see that at $K = 1$ the selection disappears and $\qstar$ collapses to a point mass, as already noted by~\citet{gladstone2026xm,shahroudi2024evaluation,bennett2026freedom}.

\noindent\textbf{Analysis in the large-$K$ case.}
At the other extreme, when $K$ is large, adding candidate mass near a target lowers its loss the most where candidates are sparse.
At $\q = \pstar$, moving a little candidate mass from a common region to a rare one therefore lowers the expected loss, so $\qstar$ gives rare regions too much mass.
As shown by~\citet{thiede2019variety} for the special case $d = 1$ and $r = 1$,
the following proposition makes this precise for general $d$ and $r$:

\begin{proposition}[Flattening law, informal; from random quantization {\citep{zador1982asymptotic,cohort2004limit}}]\label{thm:law}
When $K$ is large, we have
\[
  \qstar \;\propto\; (\pstar)^{\frac{d}{d+r}},
\]
where $d$ is the intrinsic dimension of the support of $\pstar$.
\end{proposition}

\begin{proof}[Proof sketch]
For large $K$, the expected $r$-th power of the distance from $x$ to its nearest candidate scales as $(K\,\q(x))^{-r/d}$.
The expected MoK loss is thus roughly proportional to $\int \pstar(x)\, \q(x)^{-r/d}\, dx$, and minimizing it over $\q$ gives the flattening law.
See \cref{app:coverage} for the full proof.
\end{proof}

Since $d/(d+r) < 1$, $\qstar$ is a flattened version of $\pstar$, meaning the MoK training tends to over-allocate $\q$ to rare regions.
Note that the exponent $d/(d+r)$ approaches $1$ as $d$ grows, making flattening weak for high-dimensional data such as natural images and strong for low-dimensional outputs such as actions, trajectories, and poses.

\noindent\textbf{Analysis in the medium-$K$ case.}
For general $K$ between the two extremes, exact analysis becomes difficult.
To understand this regime, following \citet{bennett2026freedom}, we consider a simplified equidistant atom model, where $\pstar$ is supported on $M$ well-separated modes with masses $m_1, \dots, m_M$, and the generated distribution puts mass $\pi_1, \dots, \pi_M$ on them.
Since only the event of hitting or missing each mode matters, the loss reduces to $\sum_{j} m_j\, (1 - \pi_j)^{K}$,
which admits the closed-form solution of \citet[Thm~7]{bennett2026freedom}:

\begin{proposition}[Optimal allocation in the atom model]\label{prop:atoms}\label{prop:atom}
For $K \ge 2$ and all $m_j > 0$, the unique optimal allocation satisfies $\pi_j = 1 - \min\bigl\{1,\; \gamma\, m_j^{-1/(K-1)}\bigr\}$, where $\gamma$ is fixed by normalization.
\end{proposition}

A derivation is given in \cref{app:atoms}. This closed form reveals that small modes can be entirely ignored, as modes with mass below the threshold $\gamma^{K-1}$ receive zero weight. As $K \to \infty$, $\pi_j \to 1/M$ regardless of the masses, recovering the flattening of \cref{thm:law} in the zero-width limit.

In summary, generically $\qstar \ne \pstar$ at every $K$.
At $K = 1$ the model collapses to a point, at small $K$ it drops the rarest modes, and as $K$ grows it over-weights rare modes and flattens the distribution.
Therefore, tuning $K$ moves between failure modes but generally cannot remove the bias.

\subsection{Limitations of target-only losses}
In this section, we further reveal that the gap between the optimal sampling distribution and $\pstar$ is not specific to the MoK loss.
\cref{prop:channels} below shows that reweighting, rescaling, or other redesigns of the MoK loss based only on target--candidate distances generally cannot close this gap.

\begin{theorem}[Exactness requires additional information]\label{prop:channels}
For any nontrivial loss that depends on the candidates only through their distances to the target,
there exists a data distribution $\pstar$ that is not even a stationary point of the expected loss.
\end{theorem}

\begin{proof}[Proof sketch]
We construct one such $\pstar$ in the equidistant atom model of \cref{prop:atoms}.
See \cref{app:necessity} for the full proof.
\end{proof}

\section{Mirrored MoK training}
\label{sec:mirror}

\noindent\textbf{Motivation.}
As shown in \cref{sec:map}, standard MoK training is biased because its loss uses only distances to the data point $x \sim \pstar$.
In particular, $\pstar$ is generally not a stationary point of MoK training, which moves $\q$ toward the MoK optimum $\qstar$ even when $\q = \pstar$.
To remove this bias, we propose to probe the generated distribution $\q$ in the same way as $\pstar$, by drawing one extra sample from the model as a second target and subtracting the resulting MoK loss from the original one.
When $\q = \pstar$, this extra sample has the same distribution as the data point, so the two distance terms have the same expected gradient and cancel.

\noindent\textbf{Construction.}
Specifically, in each step we draw one extra sample $\tilde y \sim \q$ independently of the $K$ candidates.
It is detached, so it receives no gradient and acts only as a second target.
We then apply the closest-candidate selection rule to the same $K$ candidates twice, once against the data point $x$ and once against $\tilde y$, and subtract the second distance from the first:
\begin{equation}\label{eq:mirror}
  {\setlength{\fboxsep}{5pt}\colorbox{mmoktint}{$\displaystyle\;
  \begin{gathered}
  \mathcal{L}_{\mathrm{MMoK}}
  \;=\;
  \underbrace{\mathcal{D}\bigl(\hat y_{w},\, x\bigr)}_{\text{MoK term, unchanged}}
  \;-\;
  \underbrace{\mathcal{D}\bigl(\hat y_{\tilde w},\, \tilde y\bigr)}_{\text{mirror term, against } \q},
  \\[4pt]
  \begin{aligned}
    w        &= \arg\min\nolimits_{i}\; \mathcal{D}\bigl(\hat y_i, x\bigr),\\
    \tilde w &= \arg\min\nolimits_{i}\; \mathcal{D}\bigl(\hat y_i, \tilde y\bigr).
  \end{aligned}
  \end{gathered}
  \;$}}
\end{equation}
The subtracted term mirrors the MoK term by using the same $K$ candidates, the same distance $\mathcal{D}$, and the same selection rule, with the extra sample $\tilde y$ in place of the data point $x$.
Each distance term backpropagates only through its selected candidate, so gradients reach $\hat y_{w}$ and $\hat y_{\tilde w}$ alone.
Under gradient descent, the MoK term pulls $\hat y_{w}$ toward the data point $x$, and the subtracted mirror term pushes $\hat y_{\tilde w}$ away from the extra sample $\tilde y$.

\begin{algorithm}[t]
\caption{One training step of MMoK.}
\label{alg:mirror}
\begin{algorithmic}[1]
\REQUIRE data point $x \sim \pstar$, model $\q$ with parameters $\theta$, number of candidates $K$, distance $\mathcal{D}$
\STATE Draw candidates $\hat y_{1:K} \sim \q$ i.i.d.
\STATE Draw one extra sample $\tilde y \sim \q$ and detach it \COMMENT{added}
\STATE $w \gets \arg\min_i \mathcal{D}(\hat y_i, x)$
\STATE $\tilde w \gets \arg\min_i \mathcal{D}(\hat y_i, \tilde y)$ \COMMENT{added}
\STATE $\mathcal{L}_{\mathrm{MMoK}} \gets \mathcal{D}(\hat y_w, x) - \mathcal{D}(\hat y_{\tilde w}, \tilde y)$ \COMMENT{second term added}
\STATE Update $\theta$ with $\nabla_\theta \mathcal{L}_{\mathrm{MMoK}}$, averaged over the batch
\end{algorithmic}
\end{algorithm}

\cref{alg:mirror} summarizes one training step.
MMoK adds only one extra sample per step, with no second network, no memory buffer, and no new hyperparameter.
The rest of the training loop remains unchanged.

\noindent\textbf{Theoretical guarantee.}
We first show that the data distribution $\pstar$ is a stationary point of MMoK training for any $K$ and any distance.
\begin{theorem}[Stationarity at the data distribution]\label{thm:mirror}
For a model with parameters $\theta$, any $K \ge 1$, any distance $\mathcal{D}$, and any conditioning input, if $\q = \pstar$, the expected MMoK gradient vanishes, that is, $\EE\bigl[\nabla_{\theta}\, \mathcal{L}_{\mathrm{MMoK}}\bigr] = 0$.
\end{theorem}

\begin{proof}[Proof sketch]
If $\q = \pstar$, the targets $x$ and $\tilde y$ of the two terms in \cref{eq:mirror} follow the same distribution and are independent of the candidates, so the two terms have the same expected gradient and cancel. See \cref{app:mirror_stationary} for the full proof.
\end{proof}

\cref{thm:mirror} shows that $\pstar$ is a stationary point, but leaves open whether other distributions are stationary as well.
We answer this question in the equidistant atom model of \cref{prop:atoms}, where the bias of standard MoK is explicit.
In this model, training moves mass between the modes.

\needspace{4\baselineskip}
\begin{theorem}[Unique stationary point in the atom model]\label{prop:mirror_atoms}
In the atom model of \cref{prop:atoms}, for every $K \ge 1$ and every $r > 0$, $\q = \pstar$ (that is, $\pi = m$) is the only stationary allocation of MMoK training.
\end{theorem}

\begin{proof}[Proof sketch]
In this model, the partial derivative of the expected MMoK objective in $\pi_j$, with the distribution of the extra sample held fixed, is a positive multiple of $(1 - \pi_j)^{K-1}(\pi_j - m_j)$.
Because the differences $\pi_j - m_j$ sum to zero, these partial derivatives meet the stationarity condition only at $\pi = m$, where all of them are zero.
See \cref{app:mirror_atoms} for the full proof.
\end{proof}

This result contrasts directly with \cref{prop:atoms} and shows the advantage of MMoK.
In the same model, the MoK optimum is a biased allocation that drops modes below a mass threshold and misallocates the rest, whereas the only stationary point of MMoK is the data distribution.

\section{Experiments}
\label{sec:experiments}

We evaluate MMoK in five scenarios.
A synthetic Gaussian mixture (\cref{sec:exp:toy}) verifies the theory of \cref{sec:map,sec:mirror}, and the remaining four are established MoK applications: pedestrian and vehicle trajectory prediction, tropical cyclone forecasting, robotic policy learning, and ambiguous medical image segmentation.

\subsection{Synthetic distribution experiments}
\label{sec:exp:toy}

\noindent\textbf{Settings.}
The data distribution is a three-mode isotropic Gaussian mixture in the plane ($d = 2$) with weights $(0.6, 0.3, 0.1)$ and well-separated modes.
We use small MLP generators mapping noise to the plane, trained for $100$k steps with $K \in \{2, 4, 8, 16, 20, 48\}$ candidates per step, with the same network, optimizer, and random numbers for both MoK and MMoK.
We measure convergence with the Wasserstein-2 distance ($W_2$) between $2\,000$ generator samples and $2\,000$ ground-truth samples, computed via exact optimal transport.

\begin{figure}[t]
\centering
\includegraphics[width=\columnwidth]{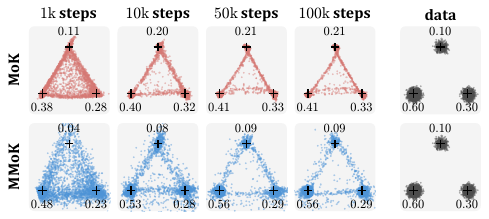}
\caption{
Generator samples at $K = 8$ during training, with MoK in the top row and MMoK in the bottom row.
Printed numbers are the mode weights, counting the samples within five standard deviations of each mode center.}
\label{fig:toy_scatter}
\end{figure}

\begin{figure}[t]
\centering
\includegraphics[width=\linewidth]{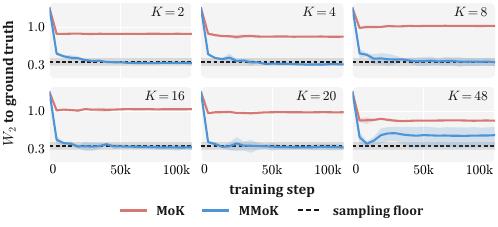}
\caption{
$W_2$ to the ground truth during training (mean $\pm$ maximum deviation over three seeds).
The dashed line marks the sampling floor, the mean $W_2$ between independent ground-truth sample sets of the same size.
}
\label{fig:toy_curves}
\end{figure}

\noindent\textbf{Results.}
\cref{fig:toy_scatter} shows the learned distributions at $K = 8$ as training progresses.
Under MoK, samples spread across the triangle formed by the three modes, with the rare mode visibly over-represented.
This confirms the over-allocation predicted by \cref{prop:atoms}, which gives the optimal allocation $(0.41, 0.35, 0.24)$ at $K = 8$.
In contrast, under MMoK, the samples concentrate around the three mode centers with nearly correct mode weights, consistent with \cref{prop:mirror_atoms}.
\cref{fig:toy_curves} quantifies this with the $W_2$ distance during training.
MoK stays far above the sampling floor throughout training, while MMoK descends steadily, approaches the floor by $25$k steps, and reaches it by the end of training for $K \le 20$.
The final MoK level is lowest at $K = 4$ and $K = 48$ and highest at $K = 8$ and $K = 16$, because the mode weights of MoK move from \cref{prop:atoms} toward \cref{thm:law} as $K$ grows.
At small $K$, the candidates often miss a mode, and the weight of the rare mode grows with $K$, as \cref{prop:atoms} predicts.
At $K = 48$, the candidates almost never miss a mode, and this weight falls back to $0.15$, close to the $0.16$ predicted by \cref{thm:law}.
At the same $K$, MMoK stays above the floor, as the weights of the two dominant modes are still slightly off.
\cref{fig:toy_w3vsk} in \cref{app:toy_details} further visually shows that the MoK bias persists at every $K$, while MMoK largely recovers the correct mode weights.
All these results are consistent with the analysis of \cref{sec:map} and show that MMoK improves upon MoK training.

\begin{widetable}[t]
\centering
\caption{
Comparison among MoK, the post-hoc power-transform correction (TB), and MMoK on ETH/UCY.
Values are means over three seeds, and the maximum deviations are in \cref{app:traj_details}.
}
\label{tab:traj}
\small
\begin{adjustbox}{max width=\textwidth}
\begin{tabular}{lccMccMccMccMccM}
\toprule
& \multicolumn{3}{c}{ADE $\downarrow$} & \multicolumn{3}{c}{FDE $\downarrow$} & \multicolumn{3}{c}{KDE-NLL $\downarrow$} & \multicolumn{3}{c}{AMD $\downarrow$} & \multicolumn{3}{c}{AMV $\downarrow$} \\
\cmidrule(lr){2-4}\cmidrule(lr){5-7}\cmidrule(lr){8-10}\cmidrule(lr){11-13}\cmidrule(lr){14-16}
scene & MoK & TB & MMoK & MoK & TB & MMoK & MoK & TB & MMoK & MoK & TB & MMoK & MoK & TB & MMoK \\
\midrule
eth   & $1.13$ & $1.38$ & $\mathbf{1.11}$ & $2.36$ & $2.87$ & $\mathbf{2.28}$ & $6.52$ & $\mathbf{3.69}$ & $4.69$ & $11.2$ & $\mathbf{2.2}$ & $8.9$ & $0.44$ & $1.56$ & $\mathbf{0.36}$ \\
hotel & $\mathbf{0.72}$ & $0.91$ & $0.74$ & $\mathbf{1.52}$ & $1.95$ & $1.54$ & $1.95$ & $\mathbf{0.98}$ & $1.20$ & $3.1$ & $\mathbf{1.5}$ & $2.2$ & $0.43$ & $1.48$ & $\mathbf{0.33}$ \\
univ  & $\mathbf{0.71}$ & $0.76$ & $\mathbf{0.71}$ & $1.55$ & $1.64$ & $\mathbf{1.52}$ & $1.97$ & $1.68$ & $\mathbf{1.30}$ & $3.4$ & $\mathbf{2.4}$ & $2.6$ & $0.32$ & $0.48$ & $\mathbf{0.22}$ \\
zara1 & $\mathbf{0.65}$ & $0.91$ & $\mathbf{0.65}$ & $1.46$ & $2.02$ & $\mathbf{1.40}$ & $0.75$ & $0.77$ & $\mathbf{0.33}$ & $2.5$ & $\mathbf{1.4}$ & $1.7$ & $0.33$ & $1.05$ & $\mathbf{0.30}$ \\
zara2 & $\mathbf{0.53}$ & $0.75$ & $0.54$ & $1.17$ & $1.66$ & $\mathbf{1.15}$ & $-0.01$ & $0.44$ & $\mathbf{-0.06}$ & $2.2$ & $\mathbf{1.1}$ & $1.6$ & $0.23$ & $0.95$ & $\mathbf{0.19}$ \\
\bottomrule
\end{tabular}
\end{adjustbox}
\end{widetable}

\subsection{Trajectory prediction}
\label{sec:exp:traj}
\label{sec:exp:ngsim}
We evaluate MMoK on trajectory prediction for both pedestrians and highway vehicles, where MoK training is widely used to capture the multimodality of future trajectories under the name ``variety loss''~\citep{gupta2018social}. In both tasks, we also compare with the post-hoc power-transform correction of~\citet{thiede2019variety} (TB), which was proposed for trajectory prediction to correct MoK predictions at inference time.

\noindent\textbf{Pedestrian trajectory prediction.}
For pedestrian trajectory prediction, we evaluate on the ETH/UCY benchmark~\citep{pellegrini2009eth,lerner2007ucy}, which contains five scenes (eth, hotel, univ, zara1, zara2).
The base pipeline is Social GAN~\citep{gupta2018social}, using its released architecture, data splits, and adversarial discriminator without modification.
The baseline loss is MoK with squared L2 distance and $K = 20$.
We report average displacement error (ADE) and final displacement error (FDE) over all $20$ generated trajectories for single-sample comparison, KDE-NLL~\citep{salzmann2020trajectron} for distributional quality, and average Mahalanobis distance (AMD) together with average maximum eigenvalue (AMV) from Social-Implicit~\citep{mohamed2022social} for distribution distance and diversity.

As shown in \cref{tab:traj}, by estimating a density from many samples for every input at inference, TB improves KDE-NLL in most scenes and AMD in all of them, but at a cost in AMV, meaning the corrected distribution is more dispersed than the original.
In contrast, MMoK generally improves on all metrics.
The decrease in KDE-NLL means the model assigns higher likelihood to the true future.
ADE is within $0.02$ of MoK in every scene and FDE is lower in four scenes, meaning that MMoK preserves the accuracy of single samples.
AMD and AMV both drop in all five scenes, meaning the predicted distribution sits closer to the ground truth.
MMoK avoids the trade-off of TB by correcting the training loss directly, so the model itself learns a closer distribution.

\noindent\textbf{Vehicle trajectory prediction.}
For vehicle trajectory prediction, we evaluate on the NGSIM US-101 dataset~\citep{fhwa2007us101}.
Given a vehicle's current trajectory consisting of positions, the task is to predict its future trajectory, consisting of positions at several future time points.
We implement the LSTM generator based on the model description of~\citet{thiede2019variety} and train with MoK and MMoK from scratch at $K\in \{8,20\}$, using the same architecture, data split, and optimization settings.
\cref{app:ngsim} provides the implementation details.
We report KDE-NLL and ADE as for pedestrians, together with the energy score~\citep{gneiting2007strictly,shahroudi2024evaluation}, which additionally measures distributional quality on whole trajectories.

\begin{table}[t]
\centering
\small
\setlength{\tabcolsep}{3pt}
\caption{Comparison among MoK, TB, and MMoK on the NGSIM US-101 dataset. The mean and maximum deviations across three seeds are reported.}
\label{tab:ngsim-us101}
\begin{adjustbox}{max width=\linewidth}
\begin{tabular}{clccc}
\toprule
$K$ & Method & KDE-NLL $\downarrow$ & Energy $\downarrow$ & ADE $\downarrow$ \\
\midrule
8 & MoK & $7.63$\std{0.34} & $4.72$\std{0.02} & $4.99$\std{0.03} \\
 & TB & $4.37$\std{0.05} & $5.73$\std{0.24} & $5.55$\std{0.06} \\
\rowcolor{mmoktint}  & MMoK & $\mathbf{2.83}$\std{0.03} & $\mathbf{4.57}$\std{0.01} & $\mathbf{4.58}$\std{0.06} \\
\midrule
20 & MoK & $3.10$\std{0.11} & $4.88$\std{0.04} & $5.51$\std{0.20} \\
 & TB & $3.08$\std{0.10} & $4.91$\std{0.09} & $5.54$\std{0.28} \\
\rowcolor{mmoktint}  & MMoK & $\mathbf{2.83}$\std{0.01} & $\mathbf{4.57}$\std{0.01} & $\mathbf{4.55}$\std{0.06} \\
\bottomrule
\end{tabular}
\end{adjustbox}
\end{table}

\begin{figure}[t]
\centering
\includegraphics[width=\columnwidth]{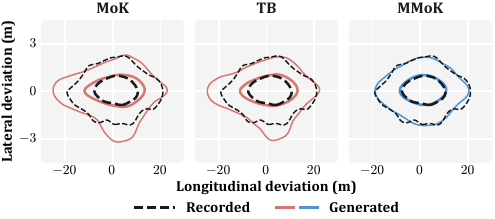}
\caption{
Endpoint distributions on NGSIM US-101 ($K = 20$), relative to the position reached at the last observed velocity. Inner and outer contours enclose $50\%$ and $90\%$ of each distribution.}
\label{fig:ngsim-k20}
\end{figure}

The quantitative results are presented in \cref{tab:ngsim-us101}.
At $K = 8$, the KDE-NLL of MoK is far above that of the other methods, meaning that some recorded futures fall where MoK assigns very low density.
On the other hand, at $K = 20$, its KDE-NLL drops, but its energy score and ADE increase.
\cref{fig:ngsim-k20} also shows that its endpoint distribution is wider than the recorded one.
This matches the analysis in \cref{sec:map} that tuning $K$ moves MoK between failure modes but does not remove its bias.
TB lowers the KDE-NLL of MoK at $K = 8$ but raises its energy score and ADE, and at $K = 20$ it changes the metrics and the endpoint distribution only slightly.
In contrast, MMoK achieves the lowest KDE-NLL, energy score, and ADE at both values of $K$, and its endpoint contours nearly coincide with the recorded ones.
Its results are also almost the same at $K = 8$ and $K = 20$, consistent with \cref{thm:mirror} that the data distribution is a stationary point for every $K$.
These results show that MMoK outperforms both MoK and TB on NGSIM at both values of $K$.

\begin{widetable}[t]
\centering
\caption{
Comparison of MoK and MMoK for tropical cyclone forecasting on TCND at the 24-hour lead.
CRPS, ensemble-mean error, and best-of-6 error are in km for track, hPa for pressure, and m/s for wind.
The mean and maximum deviations across three seeds are reported.
}
\label{tab:tc}
\small
\setlength{\tabcolsep}{2pt}
\begin{adjustbox}{max width=\textwidth}
\begin{tabular}{lccccccccccccc}
\toprule
& \multicolumn{4}{c}{\textit{Probabilistic scores}} & \multicolumn{3}{c}{\textit{Calibration}} & \multicolumn{3}{c}{\textit{Point accuracy}} & \multicolumn{3}{c}{\textit{TCNM evaluation protocol}} \\
\cmidrule(lr){2-5}\cmidrule(lr){6-8}\cmidrule(lr){9-11}\cmidrule(lr){12-14}
& \multicolumn{3}{c}{CRPS $\downarrow$} & RI BSS $\uparrow$ & \multicolumn{2}{c}{SSR $\rightarrow 1$} & Coverage $\rightarrow 0.67$ & \multicolumn{3}{c}{Ensemble-mean error $\downarrow$} & \multicolumn{3}{c}{Best-of-6 error $\downarrow$} \\
\cmidrule(lr){2-4}\cmidrule(lr){5-5}\cmidrule(lr){6-7}\cmidrule(lr){8-8}\cmidrule(lr){9-11}\cmidrule(lr){12-14}
Method & track & pressure & wind & wind & pressure & wind & track & track & pressure & wind & track & pressure & wind \\
\midrule
MoK
  & $144.5$\std{1.1} & $6.78$\std{0.20} & $4.72$\std{0.04} & $-0.02$\std{0.02}
  & $0.20$\std{0.05} & $0.21$\std{0.02} & $0.02$\std{0.01}
  & $155.7$\std{2.0} & $7.74$\std{0.07} & $5.40$\std{0.04}
  & $139.2$\std{1.8} & $5.66$\std{0.38} & $3.93$\std{0.12} \\
\rowcolor{mmoktint} MMoK
  & $\mathbf{97.0}$\std{1.5} & $\mathbf{5.03}$\std{0.11} & $\mathbf{3.43}$\std{0.05} & $\mathbf{0.16}$\std{0.01}
  & $\mathbf{0.97}$\std{0.04} & $\mathbf{0.91}$\std{0.01} & $\mathbf{0.51}$\std{0.04}
  & $\mathbf{135.5}$\std{1.5} & $\mathbf{7.04}$\std{0.16} & $\mathbf{4.81}$\std{0.09}
  & $\mathbf{86.8}$\std{1.1} & $\mathbf{3.05}$\std{0.08} & $\mathbf{2.06}$\std{0.04} \\
\bottomrule
\end{tabular}
\end{adjustbox}
\end{widetable}

\subsection{Tropical cyclone forecasting}
\label{sec:exp:tc}
\noindent\textbf{Settings.}
We evaluate MMoK on probabilistic tropical cyclone forecasting.
The goal of this task is to predict a storm's future track and intensity from its recent history and surrounding atmospheric conditions.
Note that the distributional quality is important in this task, making it a natural test for whether a model's predicted spread matches its actual error.
We use the global tropical cyclone dataset TCND~\citep{huang2025tropicyclonenet}.
Given eight past 6-hourly steps of a storm's track, intensity, and surrounding atmospheric fields, the model predicts the longitude, latitude, minimum central pressure, and maximum sustained wind over the next 24 hours.
We evaluate on the official test split, which contains storms from $2017$--$2023$.

We use TCNM~\citep{huang2025tropicyclonenet} as the baseline, whose training loss includes a MoK term with $K = 6$.
We replace only this MoK term with MMoK and train both models for $59\,063$ iterations, which is the full schedule of the released TCNM code.
For each forecast case, we draw multiple samples from the model as ensemble members.
We follow the probabilistic evaluation framework of \citet{alet2026operational} and TCBench~\citep{gomez2026tcbench} and report the metrics below.
Fair CRPS~\citep{ferro2014fair} measures distributional quality, using the great-circle distance for the track~\citep{gomez2026tcbench}.
The finite-ensemble-corrected spread-skill ratio (SSR)~\citep{fortin2014why} assesses pressure and wind calibration, with $1$ indicating calibrated spread and values below $1$ indicating under-dispersion.
The $67\%$ coverage is the fraction of cases whose observed storm center falls inside the smallest circle around the ensemble mean that contains $67\%$ of members~\citep{majumdar2010ability}.
Rapid intensification (RI) is an increase of the maximum sustained wind by at least $30$\,kt within $24$ hours~\citep{kaplan2010revised}.
Its forecast probability is the fraction of members whose predicted wind increase reaches this threshold.
We evaluate this probability with the Brier skill score (BSS), which is positive when its mean squared error is below that of always predicting the RI frequency in the training set.
We also report the ensemble-mean absolute error and, following~\citet{huang2025tropicyclonenet}, the best-of-6 error.

\noindent\textbf{Results.}
As shown in \cref{tab:tc}, the MoK ensemble is strongly under-dispersed on the test storms.
Concretely, its SSR is $0.20$ for pressure and $0.21$ for wind, and its $67\%$ circle contains the observed center in only $2\%$ of the cases.
This matches the small-$K$ analysis in \cref{sec:map}, where MoK concentrates its samples on the more frequent outcomes.
In contrast, MMoK restores the ensemble spread and improves distributional quality across all reported metrics.
It reaches an SSR of $0.97$ for pressure and $0.91$ for wind and a coverage of $0.51$, while lowering CRPS by $47.5$\,km for track, $1.75$\,hPa for pressure, and $1.29$\,m/s for wind.
The ensemble-mean error also decreases, and the best-of-6 error drops by $52.4$\,km for track, $2.61$\,hPa for pressure, and $1.87$\,m/s for wind.
All differences in CRPS, SSR, coverage, ensemble-mean error, and best-of-6 error have the same sign in each of the three seeds.
These results show that MMoK yields a better calibrated and more accurate ensemble.

\subsection{Robotic policy learning}
\label{sec:exp:robot}

\begin{table}[tb!]
\centering
\small
\setlength{\tabcolsep}{3pt}
\caption{Comparison on UR3 BlockPush with a 75:25 mode split. The mean and maximum deviations across three seeds are reported.}
\label{tab:ur3}
\begin{adjustbox}{max width=\linewidth}
\begin{tabular}{cclccc}
\toprule
$K$ & $\epsilon$ & Method & \shortstack{Mode-A\\share $\to 0.75$} & \shortstack{TV\\$\downarrow$} & \shortstack{Success\\ rate $\uparrow$} \\
\midrule
8 & 0.03 & MoK & $0.54$\std{0.02} & $0.21$\std{0.02} & $0.64$\std{0.04} \\
\rowcolor{mmoktint}  &      & MMoK & $\mathbf{0.68}$\std{0.02} & $\mathbf{0.07}$\std{0.02} & $\mathbf{0.74}$\std{0.02} \\
  & 0    & MoK & $0.57$\std{0.05} & $0.18$\std{0.05} & $0.64$\std{0.04} \\
\rowcolor{mmoktint}  &      & MMoK & $\mathbf{0.70}$\std{0.07} & $\mathbf{0.06}$\std{0.04} & $\mathbf{0.72}$\std{0.04} \\
\midrule
20 & 0.03 & MoK & $0.58$\std{0.03} & $0.17$\std{0.03} & $0.61$\std{0.07} \\
\rowcolor{mmoktint}   &      & MMoK & $\mathbf{0.67}$\std{0.00} & $\mathbf{0.08}$\std{0.00} & $\mathbf{0.69}$\std{0.03} \\
   & 0    & MoK & $0.64$\std{0.08} & $0.11$\std{0.08} & $0.55$\std{0.06} \\
\rowcolor{mmoktint}   &      & MMoK & $\mathbf{0.65}$\std{0.04} & $\mathbf{0.10}$\std{0.04} & $\mathbf{0.66}$\std{0.03} \\
\bottomrule
\end{tabular}
\end{adjustbox}
\end{table}

\paragraph{Settings.}
We evaluate MMoK on behavior cloning, where a robot learns a policy by imitating demonstrations. When demonstrations solve the same task in different ways, the policy should reproduce each way as often as the demonstrator used it. We use the BlockPush task simulated in MuJoCo~\cite{todorov2012mujoco}.
In this task, a robot arm should push a red block and a blue block that start side by side into their own goal regions.
Because the robot arm starts between the blocks and each goal is fixed, the only choice is which block to push first, giving two modes of red first (mode A) and blue first (mode B). \cref{fig:blockpush_modes} in \cref{app:ur3_examples} illustrates one successful rollout of each mode.
Because the demonstrations push the red block first $75\%$ of the time, a policy matching the data should do the same.

We build on IMLE Policy~\cite{rana2025imle}, which maps Gaussian noise to a chunk of future
actions in a single step and is trained with MoK. Its loss is the MoK objective of
\cref{eq:minofk} with $r = 1$ plus a rejection radius $\epsilon$ that excludes candidates too
close to the target from the selection, so $\epsilon = 0$ recovers \cref{eq:minofk} exactly.
We train IMLE Policy with its original loss and with MMoK at $\epsilon \in \{0, 0.03\}$ and
$K \in \{8, 20\}$, where $(0.03, 20)$ is the setting of IMLE Policy.
In MMoK, the mirror term uses the same radius around the extra sample $\tilde y$.

We report the success rate  where a rollout succeeds once each block has reached within 5\,cm of its goal.
To test whether the policy reproduces the demonstrated mode frequencies, we report the
mode-A share $\hat\pi_A$ of rollouts that push the red block first against its
target $m_A = 0.75$. We also report the total variation distance
$\mathrm{TV} = |\hat\pi_A - 0.75|$ to the data following~\citet{gibbs2002choosing}.

\begin{widefigure}[t]
\centering
\includegraphics[width=\linewidth]{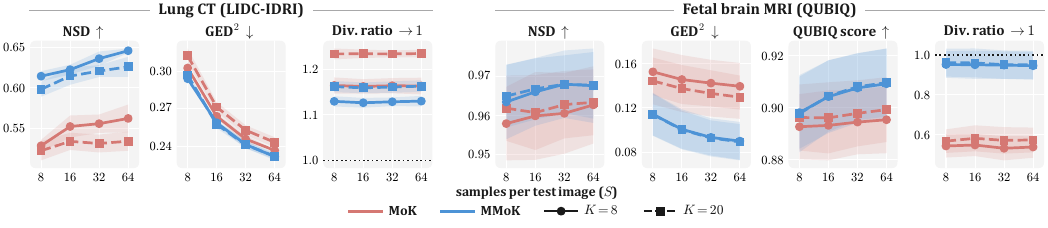}
\caption{Segmentation scores against the number $S$ of samples per test image.
Lines and bands show the mean and standard deviation over three seeds on LIDC-IDRI and over $15$ runs (five folds times three seeds) on QUBIQ.}
\label{fig:medseg}
\end{widefigure}

\paragraph{Results.}
As shown in \cref{tab:ur3}, the MoK policy over-samples the rare mode, pushing the red block first in only $54$--$64\%$ of rollouts against $75\%$ in the demonstrations, at both values of $\epsilon$.
This is the direction predicted by
\cref{prop:atoms}.
In contrast, MMoK brings the share closer to the data in every setting.
Moreover, the success rates of MMoK are higher than those of MoK in every setting.
These results show that MMoK produces a policy that better matches the demonstrated mode frequencies, while also improving task performance.

\subsection{Ambiguous medical image segmentation}
\paragraph{Settings.}

\begin{widefigure}[t]
\centering
\includegraphics[width=\linewidth]{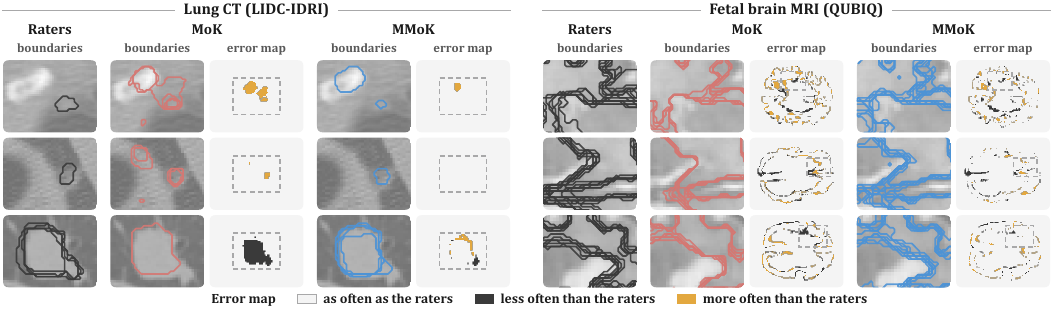}
\caption{
LIDC-IDRI (left) and QUBIQ (right) test examples at $K = 8$.
Boundaries are zoomed into the dashed region, and error maps mark the pixels where the shares of raters and of $64$ samples differ by more than $0.25$.
LIDC-IDRI shows two nodules marked by one of four raters and one marked by all four, and QUBIQ shows the images where MMoK reduces the error-pixel count the most.
}
\label{fig:qubiq_examples}
\end{widefigure}
We evaluate MMoK on medical image segmentation, which produces a binary mask labeling each pixel of a scan as foreground or background.
When the foreground is faint or has a low-contrast edge, raters draw different masks for the same image and may even disagree on its existence.
The target is then a distribution over masks, which a model should reproduce at the raters' frequencies.
We use two datasets with different kinds of ambiguity. In the LIDC-IDRI lung CT crops~\citep{armato2011lidc,kohl2018probabilistic}, four raters disagree on both whether a nodule is present and where its boundary lies. In QUBIQ brain-growth~\citep{li2405qubiq} MRI slices, seven raters agree that the foreground is present but disagree on its boundary.

We use a U-Net~\citep{ronneberger2015u} that takes the image together with a Gaussian noise image as input, so that each noise draw gives one candidate mask.
The baseline trains this model with MoK to produce $K$ candidate masks for each image, as Tyche~\citep{rakic2025tychestochasticincontextlearning} does, and we compare it with MMoK at $K \in \{8, 20\}$.
In this segmentation task, we use the sum of soft Dice and binary cross-entropy as the distance $\mathcal{D}$.
\cref{app:medseg_examples} provides the implementation details.

For each test image, we compare $S$ predicted masks with the raters' masks, varying $S \in \{8, 16, 32, 64\}$.
For accuracy, NSD measures how much of the outlines of the two majority masks (pixels marked by over half of each set) lie within 2 pixels of each other~\citep{nikolov2021clinically,maier2024metrics}.
For variation, we report GED$^2$~\citep{kohl2018probabilistic}, which compares the two mask sets as a whole and is zero only when they match in both shape and variety.
To capture this variation directly, we measure the diversity ratio, which divides the average difference between pairs of predicted masks per image by that between pairs of rater masks.
It equals 1 when the predicted masks vary as much as the rater masks.
On QUBIQ, we also report the official QUBIQ score~\citep{li2405qubiq}, which compares how often each pixel is marked in the predicted and rater masks.

\paragraph{Results.}
As shown in \cref{fig:medseg}, MMoK outperforms MoK on every metric, at every $S$ and for both values of $K$.
The clearest difference is in the diversity ratio, where MoK is far from 1 on both datasets in opposite directions, while MMoK is closer to 1 on both.
\cref{fig:qubiq_examples} shows what this means.
On LIDC-IDRI, MoK draws nodules that only one of four raters marked in most of its masks, and it often omits a nodule that all four marked.
In contrast, MMoK draws each nodule at a rate closer to the raters', so its majority mask is more accurate.
On QUBIQ, MoK draws nearly the same outline in all of its masks, failing to capture the variation in the raters' outlines.
In contrast, MMoK's outlines vary about as much as the raters', raising the diversity ratio from $0.54$ to $0.95$ at $K = 8$ and substantially lowering GED$^2$.
On both datasets, MMoK thus gives a more accurate majority mask and reproduces the disagreement among raters more closely.

\section{Conclusion}

In this paper, we analyzed the limitations of min-of-$K$ (MoK) training across all values of $K$, and showed that information beyond the distances from the candidates to the target is needed to make the generated distribution match the data distribution.
Based on this insight, we further proposed mirrored MoK (MMoK) training, which draws one extra sample from the model and subtracts the MoK loss computed against it.
We proved that, unlike MoK, MMoK training is stationary at the data distribution for any $K$, and in a simplified model, it can recover data distributions that MoK cannot.
Experimental results across five scenarios show that MMoK generally brings the generated distribution closer to the data distribution than MoK and improves task performance.
For limitations and future work, please refer to \cref{app:limitations}.

{
    \small
    \bibliographystyle{ieeenat_fullname}
    \bibliography{references}

\begin{thebibliography}{49}
\providecommand{\natexlab}[1]{#1}
\providecommand{\url}[1]{\texttt{#1}}
\expandafter\ifx\csname urlstyle\endcsname\relax
  \providecommand{\doi}[1]{doi: #1}\else
  \providecommand{\doi}{doi: \begingroup \urlstyle{rm}\Url}\fi

\bibitem[Alet et~al.(2026)Alet, Andersson, Price, Markou, El-Kadi, Masters, Li,
  Merchant, Williams, Thornton, et~al.]{alet2026operational}
Ferran Alet, Tom~R. Andersson, Ilan Price, Stratis Markou, Andrew El-Kadi,
  Dominic Masters, Amy Li, Samier Merchant, Natalie Williams, Gregory Thornton,
  et~al.
\newblock {Operational Tropical Cyclone Forecasting with {AI}}.
\newblock \emph{Nature}, 657\penalty0 (8132):\penalty0 680--688, 2026.

\bibitem[Armato~III et~al.(2011)Armato~III, McLennan, Bidaut, McNitt-Gray,
  Meyer, Reeves, Zhao, Aberle, Henschke, Hoffman, et~al.]{armato2011lidc}
Samuel~G. Armato~III, Geoffrey McLennan, Luc Bidaut, Michael~F. McNitt-Gray,
  Charles~R. Meyer, Anthony~P. Reeves, Binsheng Zhao, Denise~R. Aberle,
  Claudia~I. Henschke, Eric~A. Hoffman, et~al.
\newblock {The Lung Image Database Consortium (LIDC) and Image Database
  Resource Initiative (IDRI): A Completed Reference Database of Lung Nodules on
  CT Scans}.
\newblock \emph{Medical Physics}, 38\penalty0 (2):\penalty0 915--931, 2011.

\bibitem[Barquero et~al.(2023)Barquero, Escalera, and
  Palmero]{barquero2023belfusion}
German Barquero, Sergio Escalera, and Cristina Palmero.
\newblock {{BeLFusion}: Latent Diffusion for Behavior-Driven Human Motion
  Prediction}.
\newblock In \emph{IEEE International Conference on Computer Vision}, 2023.

\bibitem[Bennett(2026)]{bennett2026freedom}
Michael~Timothy Bennett.
\newblock {Why the Third Axis Is Freedom}.
\newblock \emph{arXiv preprint arXiv:2608.05423}, 2026.

\bibitem[Bernardo(1979)]{bernardo1979expected}
Jos{\'e}~M. Bernardo.
\newblock {Expected Information as Expected Utility}.
\newblock \emph{The Annals of Statistics}, 7\penalty0 (3):\penalty0 686--690,
  1979.

\bibitem[Bouchacourt et~al.(2016)Bouchacourt, Kumar, and
  Nowozin]{bouchacourt2016disco}
Diane Bouchacourt, M.~Pawan Kumar, and Sebastian Nowozin.
\newblock {DISCO Nets: DISsimilarity COefficient Networks}.
\newblock In \emph{Advances in Neural Information Processing Systems}, 2016.

\bibitem[Bucklew(1981)]{bucklew1981companding}
James~A. Bucklew.
\newblock {Companding and Random Quantization in Several Dimensions}.
\newblock \emph{IEEE Transactions on Information Theory}, 27\penalty0
  (2):\penalty0 207--211, 1981.

\bibitem[Cohort(2004)]{cohort2004limit}
Pierre Cohort.
\newblock {Limit Theorems for Random Normalized Distortion}.
\newblock \emph{The Annals of Applied Probability}, 14\penalty0 (1):\penalty0
  118--143, 2004.

\bibitem[Dawid et~al.(2012)Dawid, Lauritzen, and Parry]{dawid2012discrete}
A.~Philip Dawid, Steffen Lauritzen, and Matthew Parry.
\newblock {Proper Local Scoring Rules on Discrete Sample Spaces}.
\newblock \emph{The Annals of Statistics}, 40\penalty0 (1):\penalty0 593--608,
  2012.

\bibitem[Deng et~al.(2026)Deng, Li, Li, Du, and He]{deng2026drifting}
Mingyang Deng, He Li, Tianhong Li, Yilun Du, and Kaiming He.
\newblock {Generative Modeling via Drifting}.
\newblock In \emph{Advances in Neural Information Processing Systems}, 2026.

\bibitem[Fan et~al.(2017)Fan, Su, and Guibas]{fan2017pointset}
Haoqiang Fan, Hao Su, and Leonidas~J. Guibas.
\newblock {A Point Set Generation Network for {3D} Object Reconstruction from a
  Single Image}.
\newblock In \emph{IEEE Conference on Computer Vision and Pattern Recognition},
  2017.

\bibitem[{Federal Highway Administration}(2007)]{fhwa2007us101}
{Federal Highway Administration}.
\newblock {US Highway 101 Dataset}.
\newblock Fact Sheet FHWA-HRT-07-030, U.S. Department of Transportation, 2007.

\bibitem[Ferro(2014)]{ferro2014fair}
Christopher A.~T. Ferro.
\newblock {Fair Scores for Ensemble Forecasts}.
\newblock \emph{Quarterly Journal of the Royal Meteorological Society},
  140\penalty0 (683):\penalty0 1917--1923, 2014.

\bibitem[Flamary et~al.(2021)Flamary, Courty, Gramfort, Alaya, Boisbunon,
  Chambon, Chapel, Corenflos, Fatras, Fournier, et~al.]{flamary2021pot}
R{\'e}mi Flamary, Nicolas Courty, Alexandre Gramfort, Mokhtar~Z. Alaya,
  Aur{\'e}lie Boisbunon, Stanislas Chambon, Laetitia Chapel, Adrien Corenflos,
  Kilian Fatras, Nemo Fournier, et~al.
\newblock {{POT}: {Python} Optimal Transport}.
\newblock \emph{Journal of Machine Learning Research}, 22\penalty0
  (78):\penalty0 1--8, 2021.

\bibitem[Fortin et~al.(2014)Fortin, Abaza, Anctil, and Turcotte]{fortin2014why}
Vincent Fortin, Mabrouk Abaza, Fran{\c{c}}ois Anctil, and Richard Turcotte.
\newblock {Why Should Ensemble Spread Match the {RMSE} of the Ensemble Mean?}
\newblock \emph{Journal of Hydrometeorology}, 15\penalty0 (4):\penalty0
  1708--1713, 2014.

\bibitem[Gibbs and Su(2002)]{gibbs2002choosing}
Alison~L. Gibbs and Francis~Edward Su.
\newblock {On Choosing and Bounding Probability Metrics}.
\newblock \emph{International Statistical Review}, 70\penalty0 (3):\penalty0
  419--435, 2002.

\bibitem[Gladstone et~al.(2026)Gladstone, Ji, and Du]{gladstone2026xm}
Alexi Gladstone, Heng Ji, and Yilun Du.
\newblock {Explorative Modeling: Unlocking a Third Pretraining Axis and
  End-to-End Generation}.
\newblock \emph{arXiv preprint arXiv:2607.27372}, 2026.

\bibitem[Gneiting and Raftery(2007)]{gneiting2007strictly}
Tilmann Gneiting and Adrian~E. Raftery.
\newblock {Strictly Proper Scoring Rules, Prediction, and Estimation}.
\newblock \emph{Journal of the American Statistical Association}, 102\penalty0
  (477):\penalty0 359--378, 2007.

\bibitem[Gomez et~al.(2026)Gomez, S., McGraw, Tam, Azizi, Darmon, Feldmann,
  Bourdin, Poulain-{}-Auz{\'e}au, Camargo, et~al.]{gomez2026tcbench}
Milton Gomez, Saranya~Ganesh S., Marie McGraw, Frederick Iat-Hin Tam, Ilia
  Azizi, Samuel Darmon, Monika Feldmann, Stella Bourdin, Louis
  Poulain-{}-Auz{\'e}au, Suzana~J. Camargo, et~al.
\newblock {{TCBench}: A Benchmark for Tropical Cyclone Track and Intensity
  Forecasting at the Global Scale}.
\newblock \emph{arXiv preprint arXiv:2601.23268}, 2026.

\bibitem[Gupta et~al.(2018)Gupta, Johnson, Fei-Fei, Savarese, and
  Alahi]{gupta2018social}
Agrim Gupta, Justin Johnson, Li Fei-Fei, Silvio Savarese, and Alexandre Alahi.
\newblock {Social {GAN}: Socially Acceptable Trajectories with Generative
  Adversarial Networks}.
\newblock In \emph{IEEE Conference on Computer Vision and Pattern Recognition},
  2018.

\bibitem[Guzm{\'a}n-Rivera et~al.(2012)Guzm{\'a}n-Rivera, Batra, and
  Kohli]{guzman2012mcl}
Abner Guzm{\'a}n-Rivera, Dhruv Batra, and Pushmeet Kohli.
\newblock {Multiple Choice Learning: Learning to Produce Multiple Structured
  Outputs}.
\newblock In \emph{Advances in Neural Information Processing Systems}, 2012.

\bibitem[Hosseinkhani et~al.(2026)Hosseinkhani, Peng, Shramko, Aghabozorgi,
  Qian, Engst, Moazeni, Jayaraman, and Li]{hosseinkhani2026imlevla}
Kian Hosseinkhani, Qinhe Peng, George Shramko, Mehran Aghabozorgi, Jianing
  Qian, Tristan Engst, Alireza Moazeni, Dinesh Jayaraman, and Ke Li.
\newblock {{IMLE-VLA}: Fast Single-Step Action Generation for
  Vision-Language-Action Policies}.
\newblock In \emph{IEEE/RSJ International Conference on Intelligent Robots and
  Systems}, 2026.

\bibitem[Huang et~al.(2025)Huang, Mu, Zhang, Chan, Zhang, Yan, Chen, and
  Bai]{huang2025tropicyclonenet}
Cheng Huang, Pan Mu, Jinglin Zhang, Sixian Chan, Shiqi Zhang, Hanting Yan,
  Shengyong Chen, and Cong Bai.
\newblock {Benchmark Dataset and Deep Learning Method for Global Tropical
  Cyclone Forecasting}.
\newblock \emph{Nature Communications}, 16\penalty0 (1):\penalty0 5923, 2025.

\bibitem[Huang et~al.(2020)Huang, McGill, DeCastro, Fletcher, Leonard,
  Williams, and Rosman]{huang2020diversitygan}
Xin Huang, Stephen~G. McGill, Jonathan~A. DeCastro, Luke Fletcher, John~J.
  Leonard, Brian~C. Williams, and Guy Rosman.
\newblock {{DiversityGAN}: Diversity-Aware Vehicle Motion Prediction via Latent
  Semantic Sampling}.
\newblock \emph{IEEE Robotics and Automation Letters}, 5\penalty0 (4):\penalty0
  5089--5096, 2020.

\bibitem[Kaplan et~al.(2010)Kaplan, DeMaria, and Knaff]{kaplan2010revised}
John Kaplan, Mark DeMaria, and John~A. Knaff.
\newblock {A Revised Tropical Cyclone Rapid Intensification Index for the
  {A}tlantic and Eastern {N}orth {P}acific Basins}.
\newblock \emph{Weather and Forecasting}, 25\penalty0 (1):\penalty0 220--241,
  2010.

\bibitem[Kohl et~al.(2018)Kohl, Romera-Paredes, Meyer, De~Fauw, Ledsam,
  Maier-Hein, Eslami, Jimenez~Rezende, and Ronneberger]{kohl2018probabilistic}
Simon A.~A. Kohl, Bernardino Romera-Paredes, Clemens Meyer, Jeffrey De~Fauw,
  Joseph~R. Ledsam, Klaus~H. Maier-Hein, S.~M.~Ali Eslami, Danilo
  Jimenez~Rezende, and Olaf Ronneberger.
\newblock {A Probabilistic U-Net for Segmentation of Ambiguous Images}.
\newblock In \emph{Advances in Neural Information Processing Systems}, 2018.

\bibitem[Kothari and Alahi(2023)]{kothari2023safety}
Parth Kothari and Alexandre Alahi.
\newblock {Safety-Compliant Generative Adversarial Networks for Human
  Trajectory Forecasting}.
\newblock \emph{IEEE Transactions on Intelligent Transportation Systems},
  24\penalty0 (4):\penalty0 4251--4261, 2023.

\bibitem[Lee et~al.(2016)Lee, Purushwalkam Shiva~Prakash, Cogswell, Ranjan,
  Crandall, and Batra]{lee2016smcl}
Stefan Lee, Senthil Purushwalkam Shiva~Prakash, Michael Cogswell, Viresh
  Ranjan, David Crandall, and Dhruv Batra.
\newblock {Stochastic Multiple Choice Learning for Training Diverse Deep
  Ensembles}.
\newblock In \emph{Advances in Neural Information Processing Systems}, 2016.

\bibitem[Lerner et~al.(2007)Lerner, Chrysanthou, and Lischinski]{lerner2007ucy}
Alon Lerner, Yiorgos Chrysanthou, and Dani Lischinski.
\newblock {Crowds by Example}.
\newblock \emph{Computer Graphics Forum}, 26\penalty0 (3):\penalty0 655--664,
  2007.

\bibitem[Letzelter et~al.(2024)Letzelter, Perera, Rommel, Fontaine, Essid,
  Richard, and P{\'e}rez]{letzelter2024winner}
Victor Letzelter, David Perera, C{\'e}dric Rommel, Mathieu Fontaine, Slim
  Essid, Ga{\"e}l Richard, and Patrick P{\'e}rez.
\newblock {Winner-Takes-All Learners Are Geometry-Aware Conditional Density
  Estimators}.
\newblock In \emph{International Conference on Machine Learning}, 2024.

\bibitem[Li et~al.(2024)Li, Navarro, Ezhov, Bayat, Das, Kofler, Shit,
  Waldmannstetter, Paetzold, Hu, et~al.]{li2405qubiq}
Hongwei~Bran Li, Fernando Navarro, Ivan Ezhov, Amirhossein Bayat, Dhritiman
  Das, Florian Kofler, Suprosanna Shit, Diana Waldmannstetter, Johannes~C.
  Paetzold, Xiaobin Hu, et~al.
\newblock {QUBIQ: Uncertainty Quantification for Biomedical Image Segmentation
  Challenge}.
\newblock \emph{arXiv preprint arXiv:2405.18435}, 2024.

\bibitem[Maier-Hein et~al.(2024)Maier-Hein, Reinke, Godau, Tizabi, Buettner,
  Christodoulou, Glocker, Isensee, Kleesiek, Kozubek, et~al.]{maier2024metrics}
Lena Maier-Hein, Annika Reinke, Patrick Godau, Minu~D. Tizabi, Florian
  Buettner, Evangelia Christodoulou, Ben Glocker, Fabian Isensee, Jens
  Kleesiek, Michal Kozubek, et~al.
\newblock {Metrics Reloaded: Recommendations for Image Analysis Validation}.
\newblock \emph{Nature Methods}, 21\penalty0 (2):\penalty0 195--212, 2024.

\bibitem[Majumdar and Finocchio(2010)]{majumdar2010ability}
Sharanya~J. Majumdar and Peter~M. Finocchio.
\newblock {On the Ability of Global Ensemble Prediction Systems to Predict
  Tropical Cyclone Track Probabilities}.
\newblock \emph{Weather and Forecasting}, 25\penalty0 (2):\penalty0 659--680,
  2010.

\bibitem[Mohamed et~al.(2022)Mohamed, Zhu, Vu, Elhoseiny, and
  Claudel]{mohamed2022social}
Abduallah Mohamed, Deyao Zhu, Warren Vu, Mohamed Elhoseiny, and Christian
  Claudel.
\newblock {{Social-Implicit}: Rethinking Trajectory Prediction Evaluation and
  the Effectiveness of Implicit Maximum Likelihood Estimation}.
\newblock In \emph{European Conference on Computer Vision}, 2022.

\bibitem[Nikolov et~al.(2021)Nikolov, Blackwell, Zverovitch, Mendes, Livne,
  De~Fauw, Patel, Meyer, Askham, Romera-Paredes, et~al.]{nikolov2021clinically}
Stanislav Nikolov, Sam Blackwell, Alexei Zverovitch, Ruheena Mendes, Michelle
  Livne, Jeffrey De~Fauw, Yojan Patel, Clemens Meyer, Harry Askham, Bernardino
  Romera-Paredes, et~al.
\newblock {Clinically Applicable Segmentation of Head and Neck Anatomy for
  Radiotherapy: Deep Learning Algorithm Development and Validation Study}.
\newblock \emph{Journal of Medical Internet Research}, 23\penalty0
  (7):\penalty0 e26151, 2021.

\bibitem[Pellegrini et~al.(2009)Pellegrini, Ess, Schindler, and
  Van~Gool]{pellegrini2009eth}
Stefano Pellegrini, Andreas Ess, Konrad Schindler, and Luc Van~Gool.
\newblock {You'll Never Walk Alone: Modeling Social Behavior for Multi-Target
  Tracking}.
\newblock In \emph{IEEE International Conference on Computer Vision}, 2009.

\bibitem[Rakic et~al.(2024)Rakic, Wong, Gonzalez~Ortiz, Cimini, Guttag, and
  Dalca]{rakic2025tychestochasticincontextlearning}
Marianne Rakic, Hallee~E. Wong, Jose~Javier Gonzalez~Ortiz, Beth~A. Cimini,
  John~V. Guttag, and Adrian~V. Dalca.
\newblock {Tyche: Stochastic In-Context Learning for Medical Image
  Segmentation}.
\newblock In \emph{IEEE Conference on Computer Vision and Pattern Recognition},
  2024.

\bibitem[Rana et~al.(2025)Rana, Lee, Pershouse, and
  S{\"u}nderhauf]{rana2025imle}
Krishan Rana, Robert Lee, David Pershouse, and Niko S{\"u}nderhauf.
\newblock {{IMLE Policy}: Fast and Sample Efficient Visuomotor Policy Learning
  via Implicit Maximum Likelihood Estimation}.
\newblock In \emph{Robotics: Science and Systems}, 2025.

\bibitem[Ronneberger et~al.(2015)Ronneberger, Fischer, and
  Brox]{ronneberger2015u}
Olaf Ronneberger, Philipp Fischer, and Thomas Brox.
\newblock {U-Net: Convolutional Networks for Biomedical Image Segmentation}.
\newblock In \emph{International Conference on Medical Image Computing and
  Computer-Assisted Intervention}, 2015.

\bibitem[Rupprecht et~al.(2017)Rupprecht, Laina, DiPietro, Baust, Tombari,
  Navab, and Hager]{rupprecht2017learning}
Christian Rupprecht, Iro Laina, Robert DiPietro, Maximilian Baust, Federico
  Tombari, Nassir Navab, and Gregory~D. Hager.
\newblock {Learning in an Uncertain World: Representing Ambiguity through
  Multiple Hypotheses}.
\newblock In \emph{IEEE International Conference on Computer Vision}, 2017.

\bibitem[Salzmann et~al.(2020)Salzmann, Ivanovic, Chakravarty, and
  Pavone]{salzmann2020trajectron}
Tim Salzmann, Boris Ivanovic, Punarjay Chakravarty, and Marco Pavone.
\newblock {Trajectron++: Dynamically-Feasible Trajectory Forecasting with
  Heterogeneous Data}.
\newblock In \emph{European Conference on Computer Vision}, 2020.

\bibitem[Shahroudi et~al.(2024)Shahroudi, Lepson, and
  Kull]{shahroudi2024evaluation}
Novin Shahroudi, Mihkel Lepson, and Meelis Kull.
\newblock {Evaluation of Trajectory Distribution Predictions with Energy
  Score}.
\newblock In \emph{International Conference on Machine Learning}, 2024.

\bibitem[Shen and Meinshausen(2025)]{shen2025engression}
Xinwei Shen and Nicolai Meinshausen.
\newblock {Engression: Extrapolation through the Lens of Distributional
  Regression}.
\newblock \emph{Journal of the Royal Statistical Society Series B: Statistical
  Methodology}, 87\penalty0 (3):\penalty0 653--677, 2025.

\bibitem[Sz{\'e}kely and Rizzo(2013)]{szekely2013energy}
G{\'a}bor~J. Sz{\'e}kely and Maria~L. Rizzo.
\newblock {Energy Statistics: A Class of Statistics Based on Distances}.
\newblock \emph{Journal of Statistical Planning and Inference}, 143\penalty0
  (8):\penalty0 1249--1272, 2013.

\bibitem[Thiede and Brahma(2019)]{thiede2019variety}
Luca~Anthony Thiede and Pratik~Prabhanjan Brahma.
\newblock {Analyzing the Variety Loss in the Context of Probabilistic
  Trajectory Prediction}.
\newblock In \emph{IEEE International Conference on Computer Vision}, 2019.

\bibitem[Todorov et~al.(2012)Todorov, Erez, and Tassa]{todorov2012mujoco}
Emanuel Todorov, Tom Erez, and Yuval Tassa.
\newblock {MuJoCo: A Physics Engine for Model-Based Control}.
\newblock In \emph{IEEE/RSJ International Conference on Intelligent Robots and
  Systems}, 2012.

\bibitem[Vahabpour et~al.(2025)Vahabpour, Wang, Lu, Pooladzandi, and
  Roychowdhury]{vahabpour2025sog}
Arash Vahabpour, Tianyi Wang, Qiujing Lu, Omead Pooladzandi, and Vwani
  Roychowdhury.
\newblock {Diverse Imitation Learning via Self-Organizing Generative Models}.
\newblock \emph{IEEE Transactions on Neural Networks and Learning Systems},
  36\penalty0 (4):\penalty0 7145--7157, 2025.

\bibitem[Zabolotnii et~al.(2025)Zabolotnii, Muhammad, and
  Muhammad]{zabolotnii2024pedestrian}
Dmytro Zabolotnii, Yar Muhammad, and Naveed Muhammad.
\newblock {Pedestrian Motion Prediction Evaluation for Urban Autonomous
  Driving}.
\newblock In \emph{IEEE International Conference on Robotics and Biomimetics},
  2025.

\bibitem[Zador(1982)]{zador1982asymptotic}
Paul~L. Zador.
\newblock {Asymptotic Quantization Error of Continuous Signals and the
  Quantization Dimension}.
\newblock \emph{IEEE Transactions on Information Theory}, 28\penalty0
  (2):\penalty0 139--149, 1982.

\end{thebibliography}
}

\clearpage
\setcounter{page}{1}
\maketitlesupplementary
\appendix
\section{Proofs and derivations}

\subsection{Details about \cref{thm:law}}
\label{app:coverage}

\begin{assumption}[Regularity]\label{ass:regularity}
The support of $\pstar$ is a compact smooth $d$-dimensional manifold without boundary.
Both $\pstar$ and $\q$ have continuous, positive densities on it, taken with respect to its $d$-dimensional volume.
\end{assumption}

\begin{restated}{Proposition~\ref{thm:law}}{Flattening law, restated}
Under \cref{ass:regularity}, when $K$ is large, the expected MoK loss of $\q$ is approximately proportional to $\int \pstar(x)\, \q(x)^{-r/d}\, dx$.
This approximation is uniquely minimized by
\begin{align}
  \qstar \;\propto\; (\pstar)^{\frac{d}{d+r}}.
\end{align}
\end{restated}

\begin{proof}
The proof has three steps.
Step~1 computes the expected loss at a single target $x$ when $K$ is large.
Step~2 averages over targets to obtain the approximation in \cref{thm:law}.
Step~3 finds the unique minimizer of the resulting integral.

\noindent\textbf{Step 1 (Expected loss at one target).}
Fix a target $x$ and let $R = \min_i \|\hat y_i - x\|$ be the distance from $x$ to the closest candidate.
The MoK loss at this target is $R^r$, so we want $\EE[R^r]$.

\noindent\textit{Rescaling.}
When $K$ candidates share a $d$-dimensional region, each occupies a cell whose volume is about $1/K$ of the region, so the cell width, and hence $R$, shrinks like $K^{-1/d}$.
Define $T = K^{1/d}\,R$ so that $\EE[R^r] = K^{-r/d}\,\EE[T^r]$.

\noindent\textit{Probability that all candidates miss.}
The event $T > s$ means that no candidate lands within distance $s\,K^{-1/d}$ of $x$.
Let $V_d$ denote the volume of the unit ball in $\RR^d$.
The ball of radius $s\,K^{-1/d}$ around $x$ has volume about $V_d\,s^d\,K^{-1}$.
Because $\q$ is continuous and the ball is small, a single candidate falls in it with probability $\q(x)\,V_d\,s^d\,K^{-1}(1+o(1))$.
Because the candidates are independent, the probability that all $K$ miss is
\begin{align}
  \Pr[T > s]
  = \bigl(1 - \q(x)\,V_d\,s^d\,K^{-1}(1+o(1))\bigr)^{K}.
\end{align}
The limit $(1-a/K)^K \to e^{-a}$ gives $\Pr[T > s] \to \exp\bigl(-\q(x)\,V_d\,s^d\bigr)$.

\noindent\textit{Expected value of $T^r$.}
Since $T$ is nonnegative, $\EE[T^r] = \int_0^\infty r\,s^{r-1}\,\Pr[T > s]\,ds$.
Using the limit above inside this integral gives
\begin{align}
  \EE[T^r] \to \int_0^\infty r\,s^{r-1}\,e^{-\q(x)\,V_d\,s^d}\,ds.
\end{align}
Substituting $s = \q(x)^{-1/d}\,u$ replaces $s^{r-1}\,ds$ by $\q(x)^{-r/d}\,u^{r-1}\,du$ and the exponent by $-V_d\,u^d$.
The limit is therefore $C\,\q(x)^{-r/d}$, where $C = \int_0^\infty r\,u^{r-1}\,e^{-V_d\,u^d}\,du$ depends only on $d$ and $r$.
Multiplying by $K^{-r/d}$ gives
\begin{align}
  \EE[R^r] = C\,\bigl(K\,\q(x)\bigr)^{-r/d}\,\bigl(1+o(1)\bigr).
\end{align}

\noindent\textit{Why the limit can be taken inside the integral.}
Because $\q$ is bounded below by a positive constant on the compact manifold, $\Pr[T > s] \le \exp(-c\,s^d)$ for a constant $c > 0$ independent of $s$, $x$, and $K$.
Multiplied by $r\,s^{r-1}$, this bound is integrable, so dominated convergence allows taking the limit inside the integral.
For each fixed $\q$, continuity and compactness also make the $o(1)$ term uniform in $x$.

\noindent\textbf{Step 2 (Average over targets).}
In MoK training, the target is drawn from the data distribution, $x \sim \pstar$.
Averaging over $x$ shows that the expected MoK loss of $\q$ equals $C\,K^{-r/d}\,J(\q)$ up to a factor $1+o(1)$, where
\begin{align}
  J(\q) = \int \pstar(x)\,\q(x)^{-r/d}\,dx.
\end{align}
The prefactor $C\,K^{-r/d}$ does not depend on $\q$, which proves the approximation in \cref{thm:law}.

\noindent\textbf{Step 3 (Best allocation of candidates).}
We minimize $J$ over densities $\q$ with $\int \q = 1$.
Adding a little candidate density at $x$ lowers $J$ at the rate $(r/d)\,\pstar(x)\,\q(x)^{-(d+r)/d}$.
If this rate were larger at one point than at another, moving a little candidate mass between them would lower $J$.
So at the optimum the rate is the same at every $x$, which gives
\begin{align}
  \pstar(x)\,\q(x)^{-(d+r)/d} = \text{const}.
\end{align}
Solving for $\q(x)$ gives $\qstar(x) \propto \pstar(x)^{d/(d+r)}$.
Because $\q(x)^{-r/d}$ is a strictly convex function of $\q(x) > 0$, $J$ is strictly convex in $\q$.
Since no small change of the allocation lowers $J$ at $\qstar$, it is the unique minimizer.
Because $\pstar$ is continuous and positive, $\qstar$ is too, so it lies in the class of \cref{ass:regularity}.
This proves \cref{thm:law}.
\end{proof}

\noindent\textbf{Remarks.}
The calculation in the proof is classical in random quantization~\citep{zador1982asymptotic,bucklew1981companding}, and \citet[Thm.~1]{cohort2004limit} gives the rigorous statement for densities on $\RR^d$, for all $d \ge 1$ and $r > 0$.
\citet[Thm.~1]{thiede2019variety} proved that MoK is optimized by $\qstar \propto \sqrt{\pstar}$. Note that their theorem is stated for any dimension, but the proof covers only $d = 1$. A later work then cited the $\sqrt{\pstar}$ form as dimension-free~\citep{zabolotnii2024pedestrian}.
In general, the exponent is $d/(d+r)$, which equals $1/2$ only when $d = r$.

\subsection{Details about \cref{prop:atoms}}
\label{app:atoms}

\begin{assumption}[Equidistant atom model]\label{ass:atoms}
The data distribution $\pstar$ puts positive masses $m_1, \dots, m_M$ summing to $1$ on $M \ge 2$ atoms at equal pairwise distance $c > 0$.
The generated distribution $\q$ is supported on the same $M$ atoms with allocation $\pi = (\pi_1, \dots, \pi_M)$, where $\pi_j \ge 0$ and $\sum_{j=1}^M \pi_j = 1$.
\end{assumption}

\begin{restated}{Proposition~\ref{prop:atoms}}{Optimal allocation in the atom model, restated}
Under \cref{ass:atoms}, for $K \ge 2$ and all $m_j > 0$, the unique optimal allocation satisfies $\pi_j = 1 - \min\bigl\{1,\; \gamma\, m_j^{-1/(K-1)}\bigr\}$, where $\gamma$ is fixed by normalization.
\end{restated}

\begin{proof}
The proof has two steps.
Step~1 writes the expected MoK loss as a function of the allocation.
Step~2 finds the unique optimal allocation.

\noindent\textbf{Step 1 (Expected MoK loss of an allocation).}
As defined in \cref{eq:minofk}, the MoK loss is $\mathcal{L}_{\mathrm{MoK}} = \min_{i \in \{1, \dots, K\}} \mathcal{D}(\hat y_i, x)$ with $\mathcal{D}(\hat y_i, x) = \|\hat y_i - x\|^r$.
Given target atom $j$, the loss is $0$ if at least one candidate lands on atom $j$.
Otherwise all $K$ i.i.d.\ candidates land on other atoms at distance $c$, giving loss $c^r$, which happens with probability $(1 - \pi_j)^K$.
Taking the expectation over $x \sim \pstar$ gives the expected MoK loss of the allocation $\pi$,
\begin{align}
  \bar{\mathcal{L}}_{\mathrm{MoK}}(\pi) \;=\; c^r \sum_{j=1}^M m_j\, (1 - \pi_j)^K.
\end{align}

\noindent\textbf{Step 2 (Optimal allocation).}
Since $K \ge 2$ and every $m_j > 0$, each term $m_j(1 - \pi_j)^K$ is strictly convex in $\pi_j$, so $\bar{\mathcal{L}}_{\mathrm{MoK}}$ has a unique minimizer.
We find it with the Lagrangian
\begin{align}
  \Lambda(\pi, \lambda, \mu)
  &= c^r \sum_{j=1}^M m_j (1 - \pi_j)^K \nonumber \\
  &\quad + \lambda \Bigl(\sum_{j=1}^M \pi_j - 1\Bigr) - \sum_{j=1}^M \mu_j \pi_j,
\end{align}
where $\lambda$ is the multiplier of the constraint $\sum_j \pi_j = 1$ and $\mu_j \ge 0$ is the multiplier of $\pi_j \ge 0$.
Since the objective is convex and the constraints are linear, $\pi$ is the minimizer exactly when, for every $j$, $\mu_j \pi_j = 0$ and
\begin{align}
  \frac{\partial \Lambda}{\partial \pi_j}
  = -c^r K\, m_j (1 - \pi_j)^{K-1} + \lambda - \mu_j = 0.
\end{align}
If $\pi_j > 0$, then $\mu_j = 0$ and $c^r K\, m_j (1 - \pi_j)^{K-1} = \lambda$, so $1 - \pi_j = \gamma\, m_j^{-1/(K-1)}$ with $\gamma = \bigl(\lambda / (c^r K)\bigr)^{1/(K-1)}$.
If $\pi_j = 0$, then $\mu_j = \lambda - c^r K\, m_j \ge 0$, so $m_j \le \gamma^{K-1}$, that is, $\gamma\, m_j^{-1/(K-1)} \ge 1$.
The two cases combine into $1 - \pi_j = \min\bigl\{1,\, \gamma\, m_j^{-1/(K-1)}\bigr\}$, and $\gamma$ is fixed by $\sum_j \pi_j = 1$.
\end{proof}

\subsection{Details about \cref{prop:channels}}
\label{app:necessity}

\begin{definition}[Nontrivial loss]\label{def:nontrivial}
A loss that depends on the candidates only through their distances to the target is \emph{nontrivial} if, for some $M \ge 2$ points at equal pairwise distance $c > 0$ and some data distribution on these points, its expected value changes when the generated distribution moves mass among these points.
\end{definition}

\begin{restated}{Theorem~\ref{prop:channels}}{Exactness requires additional information, restated}
For any nontrivial loss that depends on the candidates only through their distances to the target, there exists a data distribution $\pstar$ that is not even a stationary point of the expected loss.
\end{restated}

\begin{proof}
For a given nontrivial loss, it suffices to exhibit one data distribution $\pstar$ for which $\pstar$ is not a stationary point of the expected loss.
We construct it in the equidistant atom model of \cref{ass:atoms} with $M \ge 3$ atoms.
Concretely, the proof has two steps.
Step~1 shows that the expected loss can be written as $\sum_j m_j \varphi(\pi_j)$, where $\varphi(\pi_j)$ is the expected loss when the target is at atom $j$.
Because the $K$ candidates are i.i.d., $\varphi$ is a polynomial of degree at most $K$.
Step~2 supposes, for contradiction, that $\pi = m$ is a stationary point of the expected loss for every $m$ with every $m_j > 0$.
With $M \ge 3$ atoms, this forces $\varphi(t)$ to equal a constant multiple of $\log t$ plus a constant.
A polynomial has this form only when it is constant, which contradicts nontriviality.

\noindent\textbf{Step 1 (Reduction to a polynomial).}
Consider $M \ge 2$ atoms at equal pairwise distance $c > 0$.
Every target--candidate distance is then $0$ (hit) or $c$ (miss).
Since the loss reads only these distances, its average over any randomization is a function $\ell(b)$ of the hit pattern $b \in \{0,1\}^K$, where $b_i = \mathbf{1}\{\hat y_i = x\}$.
When the target is at atom $j$, the entries of $b$ are i.i.d.\ $\mathrm{Bern}(\pi_j)$, so the expected loss is $\varphi(\pi_j)$, where
\begin{align}
  \varphi(t) = \sum_{b \in \{0,1\}^K} \ell(b) \prod_{i=1}^K t^{b_i} (1-t)^{1-b_i}
\end{align}
is a polynomial in $t$ of degree at most $K$.
Averaging over the target atom gives $\bar{\mathcal{L}}(\pi) = \sum_{j=1}^M m_j\, \varphi(\pi_j)$.
This holds with the same $\varphi$ for every $M$, because $\varphi$ depends only on $c$ and the loss.
If the loss is nontrivial (\cref{def:nontrivial}), then $\varphi$ is not constant for some $c > 0$, because a constant $\varphi$ makes $\bar{\mathcal{L}}$ independent of the allocation.

\noindent\textbf{Step 2 (Impossibility for $M \ge 3$).}
Fix a distance $c > 0$ for which $\varphi$ is not constant.
Take $M = 3$ atoms at pairwise distance $c$, for example the vertices of an equilateral triangle in a plane.
Suppose that $\pi = m$ is a stationary point of $\bar{\mathcal{L}}$ over allocations for every mass vector $m$ with every $m_j > 0$.
Let $\psi(t) = t\, \varphi'(t)$.
Since $\partial \bar{\mathcal{L}} / \partial \pi_j = m_j \varphi'(\pi_j)$ and every $m_j > 0$, the first-order condition at $\pi = m$ requires $\psi(m_j)$ to be the same for all $j$.
For any $u, v \in (0,1)$, take $w > 0$ with $u + w < 1$ and $v + w < 1$.
The mass vectors $(u, w, 1-u-w)$ and $(v, w, 1-v-w)$ then give $\psi(u) = \psi(w) = \psi(v)$.
Hence $\psi$ equals a constant $a$ on $(0,1)$, so $\varphi'(t) = a/t$ and $\varphi(t) = a \log t + \mathrm{const}$.
This follows the argument of \citet{bernardo1979expected} in the discrete form of \citet[Sec.~3.2]{dawid2012discrete}.
Because $\varphi$ is a polynomial, $a = 0$, so $\varphi$ is constant, contradicting the choice of $c$.

Therefore, for some $m$ with every $m_j > 0$, the allocation $\pi = m$ is not a stationary point of $\bar{\mathcal{L}}$.
It indicates that the data distribution $\pstar$ with these masses is not a stationary point of the expected loss, so it is not a minimizer either.
\end{proof}

\subsection{Details about \cref{thm:mirror}}
\label{app:mirror_stationary}

\begin{restated}{Theorem~\ref{thm:mirror}}{Stationarity at the data distribution, restated}
For a model with parameters $\theta$, any $K \ge 1$, any distance $\mathcal{D}$, and any conditioning input, if $\q = \pstar$, the expected MMoK gradient vanishes, i.e., $\EE\bigl[\nabla_{\theta}\, \mathcal{L}_{\mathrm{MMoK}}\bigr] = 0$.
\end{restated}

\begin{proof}
The proof has two steps.
Step~1 writes the gradient of $\mathcal{L}_{\mathrm{MMoK}}$ as a difference of two terms.
Step~2 shows that the two terms have the same expectation when $\q = \pstar$.

\noindent\textbf{Step 1 (The gradient as a difference of two terms).}
For a target $z$, let $w(z) = \arg\min_i \mathcal{D}(\hat y_i, z)$, with ties broken by a fixed rule, so that $w = w(x)$ and $\tilde w = w(\tilde y)$ in \cref{eq:mirror}.
Each term of \cref{eq:mirror} backpropagates only through its selected candidate, and $\tilde y$ is detached, so
\begin{align}
  \nabla_\theta\, \mathcal{L}_{\mathrm{MMoK}}
  \;=\;
  g(x)
  \;-\;
  g(\tilde y),
\end{align}
where $g(z) = \nabla_\theta\, \mathcal{D}\bigl(\hat y_{w(z)}, z\bigr)$ is the gradient of the selected distance for the current $K$ candidates, with the target $z$ held fixed.

\noindent\textbf{Step 2 (The two terms have the same expectation).}
Both the data point $x \sim \pstar$ and the extra sample $\tilde y \sim \q$ are independent of the $K$ candidates.
Given the $K$ candidates, the first term then has expectation $\EE_{z \sim \pstar}\, g(z)$ and the second term has expectation $\EE_{z \sim \q}\, g(z)$.
When $\q = \pstar$, these two expectations are equal for every draw of the candidates, so averaging over the candidates gives $\EE\bigl[\nabla_\theta\, \mathcal{L}_{\mathrm{MMoK}}\bigr] = 0$.
\end{proof}

\subsection{Details about \cref{prop:mirror_atoms}}
\label{app:mirror_atoms}

\begin{restated}{Theorem~\ref{prop:mirror_atoms}}{Unique stationary point in the atom model, restated}
Under \cref{ass:atoms}, for every $K \ge 1$ and every $r > 0$, $\q = \pstar$ (that is, $\pi = m$) is the only stationary allocation of MMoK training.
\end{restated}

\begin{proof}
The proof has three steps.
Step~1 derives the expected MMoK objective and its gradient as functions of the allocation.
Step~2 shows that $\pi = m$ is the unique stationary point with every $\pi_j > 0$.
Step~3 shows that no stationary point has a mode without mass.

\noindent\textbf{Step 1 (MMoK objective and its gradient).}
As shown in \cref{app:atoms}, the expected MoK term of \cref{eq:mirror} is $c^r \sum_j m_j (1 - \pi_j)^K$.
The mirror term has the same form with the extra sample drawn from $\q$ in place of the data point, so its mode probabilities are $\tilde\pi = \mathrm{sg}[\pi]$, where $\mathrm{sg}[\cdot]$ denotes the stop-gradient operator.
So the expected MMoK objective of an allocation is:
\begin{align}
  \bar{\mathcal{L}}_{\mathrm{MMoK}}(\pi; \tilde\pi)
  \;=\;
  c^r \sum_{j=1}^{M} (m_j - \tilde\pi_j)(1 - \pi_j)^K.
\end{align}
We write $\partial_j \bar{\mathcal{L}}_{\mathrm{MMoK}}$ for its partial derivative in $\pi_j$, which is
\begin{align}
  \partial_j \bar{\mathcal{L}}_{\mathrm{MMoK}}
  &= \frac{\partial \bar{\mathcal{L}}_{\mathrm{MMoK}}}{\partial \pi_j}\bigg|_{\tilde\pi = \pi} \nonumber \\
  &= -c^r K (1 - \pi_j)^{K-1}(m_j - \pi_j).
\end{align}
For $K = 1$, the factor $(1 - \pi_j)^{K-1}$ equals $1$, also at $\pi_j = 1$.

\noindent\textbf{Step 2 (Uniqueness).}
An allocation is a stationary point of MMoK training when, for some constant $a$, $\partial_j \bar{\mathcal{L}}_{\mathrm{MMoK}} = a$ for every $j$ with $\pi_j > 0$ and $\partial_j \bar{\mathcal{L}}_{\mathrm{MMoK}} \ge a$ for every $j$ with $\pi_j = 0$.
These conditions have the same form as those in Step~2 of \cref{app:atoms}, here with $\tilde\pi$ held fixed.
This condition concerns the allocation itself.
A parameterization such as a softmax over logits moves mass into a mode at a rate proportional to the mass that mode already has, so training can slow down near an allocation that drops a mode, although such an allocation is not stationary.
For an allocation with every $\pi_j > 0$, the condition says that all $\partial_j \bar{\mathcal{L}}_{\mathrm{MMoK}}$ are equal to $a$.
Since $M \ge 2$, every $\pi_j < 1$, so $c^r K (1 - \pi_j)^{K-1} > 0$ and $\partial_j \bar{\mathcal{L}}_{\mathrm{MMoK}}$ has the same sign as $\pi_j - m_j$.
If $a > 0$, then $\pi_j > m_j$ for all $j$, and summing gives $1 > 1$, a contradiction.
Similarly, $a < 0$ gives $1 < 1$.
Thus $a = 0$ and $\pi_j = m_j$ for all $j$, so $\pi = m$ is the unique stationary point with every $\pi_j > 0$.

\noindent\textbf{Step 3 (No mode without mass).}
Suppose that a stationary point has $\pi_j = 0$ for some $j$.
Then $\partial_j \bar{\mathcal{L}}_{\mathrm{MMoK}} = -c^r K m_j < 0$.
The modes with mass have total allocation $1$ but total data mass less than $1$, so one of them, say $k$, has $\pi_k > m_k$, and then $\partial_k \bar{\mathcal{L}}_{\mathrm{MMoK}} \ge 0$.
Stationarity gives $a = \partial_k \bar{\mathcal{L}}_{\mathrm{MMoK}} \ge 0$ and $\partial_j \bar{\mathcal{L}}_{\mathrm{MMoK}} \ge a \ge 0$, a contradiction.
Hence every stationary point has every $\pi_j > 0$, and by Step~2 it is $\pi = m$.
\end{proof}

\section{Experimental details and additional results}
\label{app:exp_details}

\cref{tab:exp_settings} summarizes the training and evaluation settings, and the subsections below give the remaining details and additional results.
In every experiment, MMoK draws the extra sample from the current model with its own noise, as in \cref{alg:mirror}.

\begin{widetable}[t]
\centering
\caption{Training and evaluation settings of the synthetic, ETH/UCY, NGSIM, tropical cyclone, robotics, and segmentation experiments.
MoK and MMoK use the same settings within each experiment.
The segmentation column shows LIDC-IDRI\,/\,QUBIQ values where they differ.}
\label{tab:exp_settings}
\small
\begin{adjustbox}{max width=\textwidth}
\begin{tabular}{@{}lllllll@{}}
\toprule
& Synthetic & ETH/UCY & NGSIM & Cyclone & Robotics & Segmentation \\
\midrule
\multicolumn{7}{@{}l}{\textit{Model}} \\ \addlinespace[2pt]
Network
  & \begin{tabular}[t]{@{}l@{}}MLP (2${\times}$64,\\SiLU)\end{tabular}
  & \begin{tabular}[t]{@{}l@{}}Social GAN\\(no pooling)\end{tabular}
  & \begin{tabular}[t]{@{}l@{}}stochastic LSTM\\(hidden size $128$,\\$12$-D latent,\\$0.22$M parameters)\end{tabular}
  & TCNM
  & \begin{tabular}[t]{@{}l@{}}IMLE Policy 1-D\\convolutional U-Net\\($61.2$M parameters)\end{tabular}
  & \begin{tabular}[t]{@{}l@{}}U-Net, 4 levels (32,\\64, 128, 256 channels)\end{tabular}
  \\ \addlinespace[2pt]
\midrule
\multicolumn{7}{@{}l}{\textit{Loss}} \\ \addlinespace[2pt]
$K$
  & $\{2, 4, 8, 16, 20, 48\}$
  & $20$
  & $\{8, 20\}$
  & $6$
  & $\{8, 20\}$
  & $\{8, 20\}$
  \\ \addlinespace[2pt]
Distance, $r$
  & Euclidean, $r{=}1$
  & \begin{tabular}[t]{@{}l@{}}squared Euclidean,\\$r{=}2$\end{tabular}
  & \begin{tabular}[t]{@{}l@{}}squared Euclidean,\\lateral weight $20$,\\$r{=}2$\end{tabular}
  & \begin{tabular}[t]{@{}l@{}}squared Euclidean\\(joint), $r{=}2$\end{tabular}
  & Euclidean, $r{=}1$
  & \begin{tabular}[t]{@{}l@{}}soft Dice + pixel-wise\\binary cross-entropy\\(equal weights)\end{tabular}
  \\ \addlinespace[2pt]
\begin{tabular}[t]{@{}l@{}}MMoK extra\\sample\end{tabular}
  & \begin{tabular}[t]{@{}l@{}}$(K{+}1)$-th output of\\same forward pass\end{tabular}
  & \begin{tabular}[t]{@{}l@{}}1 extra forward\\pass, fresh noise\end{tabular}
  & \begin{tabular}[t]{@{}l@{}}1 extra forward\\pass, fresh noise\end{tabular}
  & \begin{tabular}[t]{@{}l@{}}same forward pass,\\new random decoder route,\\fresh Gaussian noise\end{tabular}
  & \begin{tabular}[t]{@{}l@{}}1 extra forward pass,\\non-EMA weights\end{tabular}
  & \begin{tabular}[t]{@{}l@{}}$(K{+}1)$-th output of\\same forward pass\end{tabular}
  \\ \addlinespace[2pt]
\midrule
\multicolumn{7}{@{}l}{\textit{Optimization}} \\ \addlinespace[2pt]
Optimizer
  & Adam
  & \begin{tabular}[t]{@{}l@{}}Adam, generator\\gradient-norm\\clipping $2.0$\end{tabular}
  & \begin{tabular}[t]{@{}l@{}}Adam, gradient-norm\\clipping $10.0$\end{tabular}
  & \begin{tabular}[t]{@{}l@{}}AdamW,\\$(\beta_1, \beta_2){=}(0.5, 0.999)$,\\weight decay $0.01$\end{tabular}
  & \begin{tabular}[t]{@{}l@{}}AdamW, weight decay\\$10^{-6}$, gradient-norm\\clipping $1.0$\end{tabular}
  & Adam
  \\ \addlinespace[2pt]
\begin{tabular}[t]{@{}l@{}}Learning rate,\\schedule\end{tabular}
  & \begin{tabular}[t]{@{}l@{}}$2{\times}10^{-3}$, cosine\\${\to}\,10^{-4}$ ($60$k steps),\\then constant ($40$k)\end{tabular}
  & \begin{tabular}[t]{@{}l@{}}$10^{-4}$ (generator),\\$10^{-3}$ (discriminator),\\constant\end{tabular}
  & \begin{tabular}[t]{@{}l@{}}$10^{-3}$, ${\times}0.3$ at\\$4$k and $7$k steps\end{tabular}
  & $10^{-4}$, cosine
  & \begin{tabular}[t]{@{}l@{}}$10^{-4}$, cosine\\+\,$500$ warm-up steps\end{tabular}
  & $10^{-4}$, constant
  \\ \addlinespace[2pt]
Batch size
  & $96$
  & $64$
  & $256$
  & $96$
  & $64$
  & $16$
  \\ \addlinespace[2pt]
\begin{tabular}[t]{@{}l@{}}Training\\length\end{tabular}
  & $100$k steps
  & \begin{tabular}[t]{@{}l@{}}$5\,540$--$7\,818$\\iterations (per scene)\end{tabular}
  & $10$k steps
  & \begin{tabular}[t]{@{}l@{}}$59\,063$ iterations\\($101$ epochs)\end{tabular}
  & \begin{tabular}[t]{@{}l@{}}$345\,450$ steps\\($150$ epochs)\end{tabular}
  & $30$k\,/\,$20$k steps
  \\ \addlinespace[2pt]
EMA
  & --
  & --
  & --
  & --
  & \begin{tabular}[t]{@{}l@{}}IMLE Policy\\schedule\end{tabular}
  & $0.999$
  \\ \addlinespace[2pt]
\midrule
\multicolumn{7}{@{}l}{\textit{Evaluation}} \\ \addlinespace[2pt]
\begin{tabular}[t]{@{}l@{}}Samples\\per input\end{tabular}
  & \begin{tabular}[t]{@{}l@{}}$30\,000$ (mode weights)\\$10\,000$ ($W_2$)\end{tabular}
  & \begin{tabular}[t]{@{}l@{}}$500$ (KDE-NLL)\\$20$ (other)\end{tabular}
  & \begin{tabular}[t]{@{}l@{}}$500$ (KDE-NLL)\\$64$ (energy score)\\$20$ (ADE)\end{tabular}
  & $51$
  & \begin{tabular}[t]{@{}l@{}}$200$ rollouts\\per checkpoint\end{tabular}
  & $8$, $16$, $32$, $64$
  \\ \addlinespace[2pt]
Checkpoint
  & final ($100$k)
  & last saved
  & \begin{tabular}[t]{@{}l@{}}best validation\\weighted energy score\\(every $500$ steps)\end{tabular}
  & final
  & \begin{tabular}[t]{@{}l@{}}metrics averaged over\\epochs $100$, $125$, $150$\\(EMA)\end{tabular}
  & \begin{tabular}[t]{@{}l@{}}LIDC-IDRI by validation\\GED$^2$, QUBIQ average\\of $18$--$20$k (both EMA)\end{tabular}
  \\
\bottomrule
\end{tabular}
\end{adjustbox}
\end{widetable}

\subsection{Synthetic distribution experiments}
\label{app:toy_details}

\noindent\textbf{Experimental details}\par
\noindent\textit{Data and network.}
The data distribution $\pstar$ is a mixture of three isotropic Gaussians in $\RR^2$ with centers $(0, 0)$, $(3, 0)$, and $(1.5, 2.6)$, standard deviation $\sigma = 0.15$, and weights $(0.6, 0.3, 0.1)$.
The network is an MLP that accepts $8$-dimensional standard Gaussian noise as input.

\noindent\textit{Evaluation.}
For the mode weights in \cref{fig:toy_scatter,fig:toy_w3vsk}, each model sample is assigned to its nearest center if it lies within $5\sigma$ of that center.
Each $W_2$ value is the mean over $5$ disjoint blocks of $2\,000$ samples taken from $10\,000$ samples of each distribution, computed with the POT library~\citep{flamary2021pot}.
The ground-truth samples are the same for all comparisons.

\smallskip\noindent\textbf{Additional results}

\begin{widefigure}[t]
\centering
\includegraphics[width=\linewidth]{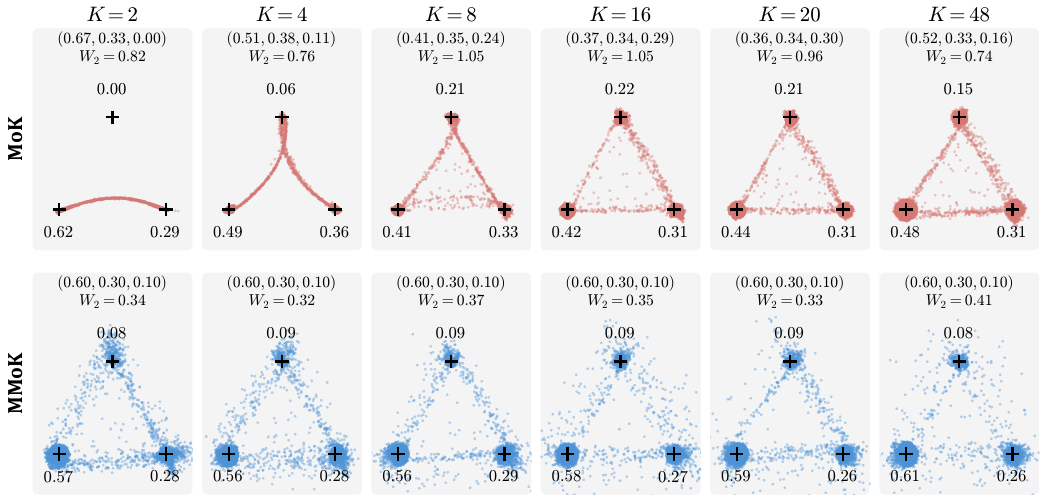}
\caption{Learned distributions of MoK and MMoK after $100$k training steps.
Each panel prints the measured mode weights, the predicted weights (largest, middle, rare mode), and the $W_2$ to the ground truth.
The predictions are from \cref{prop:atoms} for MoK at $K \le 20$, from \cref{thm:law} for MoK at $K = 48$, and from \cref{prop:mirror_atoms} for MMoK.}
\label{fig:toy_w3vsk}
\end{widefigure}
\noindent\cref{fig:toy_w3vsk} shows that the MoK bias persists at every $K$, while MMoK largely recovers the correct mode weights.

\subsection{Trajectory prediction}
\label{app:traj_details}
\label{app:ngsim}

\noindent\textbf{Experimental details}\par
\noindent\textit{ETH/UCY training.}
As in Social GAN~\citep{gupta2018social}, each minimum over $K$ in MoK and MMoK is taken per time window over the distance summed across all pedestrians in that window.
All other training settings follow the released Social GAN models.

\noindent\textit{ETH/UCY metrics.}
ADE and FDE in \cref{tab:traj} average the displacement error over all $20$ generated trajectories.
KDE-NLL follows~\citet{salzmann2020trajectron} with a per-step cap of $20$.

\noindent\textit{TB correction in ETH/UCY.}
At each future step $t$, TB~\citep{thiede2019variety} raises the per-step KDE $\hat q_t$ used for KDE-NLL to a power $\bar k$ and renormalizes.
KDE-NLL for TB is evaluated directly on the transformed density $\hat q_t^{\bar k} / Z_t$.
AMD and AMV use $20$ draws per step from the transformed density itself, obtained by rejection sampling.
ADE and FDE need whole trajectories, which TB does not define, so $20$ of the $500$ generated trajectories $y = (y_1, \dots, y_{12})$ are resampled with replacement using the geometric mean of their $12$ per-step weights $\hat q_t(y_t)^{\bar k} / \hat q_t(y_t) = \hat q_t(y_t)^{\bar k - 1}$,
\begin{align}
  w(y) \;\propto\; \prod_{t=1}^{12} \hat q_t(y_t)^{(\bar k - 1)/12}.
\end{align}
The exponent $\bar k$ is selected per checkpoint by validation KDE-NLL on a grid of $78$ values spanning $[0.001, 3]$, more than the $25$ values of the published grid of~\citet[Sec.~7.1]{thiede2019variety} over the same range.

\noindent\textit{NGSIM data.}
The NGSIM US-101 dataset~\citep{fhwa2007us101} contains three recording periods, which are split by vehicle--period group into $159\,570$ training, $33\,945$ validation, and $35\,039$ test windows.
Each window contains longitudinal and lateral positions at $-3.2$, $-1.6$, and $0$\,s as input and at $1.6$, $3.2$, and $4.8$\,s as target, in meters relative to the last observed position.

\noindent\textit{NGSIM metrics.}
KDE-NLL follows the same evaluation as ETH/UCY, using the public Trajectron++ code~\citep{salzmann2020trajectron}.\footnote{\raggedright\url{https://github.com/StanfordASL/Trajectron-plus-plus/blob/1031c7bd1a444273af378c1ec1dcca907ba59830/trajectron/evaluation/evaluation.py}}
All metrics use unweighted positions in meters.
We estimate the energy score~\citep{shahroudi2024evaluation} from pairs of distinct samples.
Similar to ETH/UCY, ADE averages over all $20$ generated trajectories.

\noindent\textit{NGSIM TB correction.}
TB follows the ETH/UCY procedure, with a separate exponent $\bar k_t$ for each future step:
\begin{align}
  q^{\mathrm{TB}}_t(y_t) = \frac{\hat q_t(y_t)^{\bar k_t}}{Z_t}.
\end{align}
Each $\bar k_t$ is selected by validation KDE-NLL on a fixed $2\,048$-window validation subset, using a grid of $49$ values in $[0.001, 3]$.
For trajectory metrics, $64$ trajectories are resampled from the $500$ generated trajectories using weights
\begin{align}
  w(y) \;\propto\; \prod_{t=1}^{3} \hat q_t(y_t)^{(\bar k_t - 1)/3}.
\end{align}
KDE-NLL evaluates the per-step correction of TB, and the energy score and ADE evaluate its trajectory-level resampling, as for ETH/UCY.

\noindent\textit{NGSIM visualization.}
\cref{fig:ngsim-k20} uses $16$ generated trajectories at $K = 20$ for each of $919$ test histories, one chosen at random from each test vehicle--period group.
It shows positions at $4.8$\,s relative to where each vehicle would arrive at its last observed speed and direction.
Each distribution is a smoothed $2$-D histogram of the pooled endpoints.
For TB, the endpoints come from the first $16$ of the $64$ resampled trajectories.

\smallskip\noindent\textbf{Additional results}\par
\noindent \cref{tab:traj_detail} reports the max-deviation over three seeds for each entry of \cref{tab:traj}.

\begin{widetable}[t]
\centering
\caption{Maximum deviations over three seeds for the ETH/UCY results in \cref{tab:traj}.
Each entry is the largest absolute difference between a single seed and the mean of the three seeds, with the same rows and columns as \cref{tab:traj}.}
\label{tab:traj_detail}
\small
\begin{adjustbox}{max width=\textwidth}
\begin{tabular}{lccMccMccMccMccM}
\toprule
& \multicolumn{3}{c}{ADE} & \multicolumn{3}{c}{FDE} & \multicolumn{3}{c}{KDE-NLL} & \multicolumn{3}{c}{AMD} & \multicolumn{3}{c}{AMV} \\
\cmidrule(lr){2-4}\cmidrule(lr){5-7}\cmidrule(lr){8-10}\cmidrule(lr){11-13}\cmidrule(lr){14-16}
scene & MoK & TB & MMoK & MoK & TB & MMoK & MoK & TB & MMoK & MoK & TB & MMoK & MoK & TB & MMoK \\
\midrule
eth   & $0.04$ & $0.06$ & $0.00$ & $0.08$ & $0.15$ & $0.05$ & $0.39$ & $0.27$ & $0.69$ & $5.4$ & $0.2$ & $9.7$ & $0.08$ & $0.33$ & $0.07$ \\
hotel & $0.02$ & $0.05$ & $0.03$ & $0.03$ & $0.12$ & $0.06$ & $0.60$ & $0.18$ & $0.43$ & $0.4$ & $0.1$ & $0.3$ & $0.04$ & $0.32$ & $0.06$ \\
univ  & $0.03$ & $0.03$ & $0.03$ & $0.08$ & $0.07$ & $0.07$ & $0.43$ & $0.29$ & $0.21$ & $0.2$ & $0.2$ & $0.2$ & $0.06$ & $0.09$ & $0.03$ \\
zara1 & $0.01$ & $0.06$ & $0.01$ & $0.03$ & $0.10$ & $0.04$ & $0.05$ & $0.14$ & $0.06$ & $0.1$ & $0.0$ & $0.1$ & $0.03$ & $0.18$ & $0.02$ \\
zara2 & $0.01$ & $0.01$ & $0.02$ & $0.02$ & $0.03$ & $0.06$ & $0.17$ & $0.08$ & $0.03$ & $0.0$ & $0.1$ & $0.1$ & $0.00$ & $0.07$ & $0.02$ \\
\bottomrule
\end{tabular}
\end{adjustbox}
\end{widetable}

\subsection{Tropical cyclone forecasting}
\label{app:tc_details}

\noindent\textbf{Experimental details}\par

\noindent\textit{Model and loss.}
In the released TCNM code~\citep{huang2025tropicyclonenet}, the generator loss is the sum of the MoK term, an adversarial term, and an auxiliary term.
MMoK replaces only the MoK term with the MMoK loss.

\noindent\textit{Training.}
Training follows the released TCNM code~\citep{huang2025tropicyclonenet} except for a routing bug fix, deterministic cuDNN, a per-seed data order, and skipping non-finite generator steps, all applied to both MoK and MMoK.
Both methods are trained for the full schedule of the released TCNM code ($101$ epochs) and evaluated at the final checkpoint.

\noindent\textit{Evaluation.}
The released test split of TCND~\citep{huang2025tropicyclonenet} contains $10\,744$ windows, each with $8$ input steps and $4$ target steps.
Of these, $10\,478$ have time stamps spaced by exactly $6$\,h and are used for evaluation.
Apart from best-of-6, all metrics are computed on ensembles of $50$ members.
For the $67\%$ coverage, we draw one more member per window, hold out each of the $51$ members in turn, and build the circle from the other $50$.
CRPS, SSR, ensemble-mean error, and RI use the first $50$ members.
Best-of-6 uses the first $48$ members in $8$ disjoint groups of $6$ and averages the per-variable minimum error over the groups.
The RI evaluation rounds both predicted and observed winds to the storm's reporting grid, for example $5$\,kt steps or integer m/s.
As a result, the effective threshold is $15.3$\,m/s, which separates $25$\,kt from $30$\,kt $= 15.43$\,m/s on the $5$\,kt grid.
Following~\citet{kaplan2010revised}, BSS uses as reference the RI frequency in the training data, which is $3.685\%$ over the $54\,005$ training windows whose time stamps are spaced by exactly $6$\,h.

\smallskip\noindent\textbf{Additional results}\par
\noindent The released TCNM model~\citep{huang2025tropicyclonenet} is the checkpoint after iteration $16\,000$ of the training schedule, not the final one.
At that iteration, our MoK model matches the released model, with a 24-hour SSR of $0.78$ for pressure and $0.80$ for wind, compared to $0.77$ and $0.81$ for the released model.
However, with further training, the MoK ensemble keeps narrowing, and its SSR drops to $0.20$ and $0.21$ at the final checkpoint.
MMoK at the final checkpoint is better calibrated than MoK at iteration $16\,000$, with an SSR of $0.97$ and $0.91$ and a $67\%$ coverage of $0.51$, against $0.78$, $0.80$, and $0.30$.
Also, its track CRPS is lower, $97.0$ against $105.7$\,km, and its pressure and wind CRPS are close, $5.03$ against $5.09$\,hPa and $3.43$ against $3.46$\,m/s.

\subsection{Robotic policy learning}
\label{app:ur3_examples}

\noindent\textbf{Experimental details}\par

\noindent\textit{Data.}
To evaluate MoK and MMoK under an imbalanced setting, we draw $240$ mode-A and $80$ mode-B demonstrations at random as the training set from the demonstrations released with IMLE Policy~\citep{rana2025imle}, giving $320$ demonstrations at a $75{:}25$ ratio.

\noindent\textit{MMoK term.}
As mentioned in \cref{sec:exp:robot}, IMLE Policy uses the MoK loss with an additional rejection radius $\epsilon$ that excludes the candidates within $\epsilon$ of the target.
The mirror term applies the same rejection radius around the extra sample $\tilde y$.
We use the same $\epsilon$ in both terms of MMoK and in the MoK baseline, namely $0.03$ as in IMLE Policy and $0$ for no rejection.

\noindent\textit{Evaluation.}
During evaluation, all rollouts use the same $200$ initial states across checkpoints and methods.
A rollout lasts at most $1000$ environment steps and ends once both blocks reach their goals.
In the estimation of the mode-A share $\hat{\pi_A}$, each rollout is marked as mode A or B according to which block first moves more than $1$\,cm from its starting position.

\noindent\textit{Implementation adjustments.}
Compared with the released IMLE Policy configuration, we train both MoK and MMoK for $150$ instead of $500$ epochs to keep the total cost of the training runs manageable.
For a more precise estimate of the mode-A share, each checkpoint is evaluated with $200$ instead of $50$ rollouts.

\smallskip\noindent\textbf{Additional results}\par
\noindent \cref{fig:blockpush_modes} shows one successful rollout of each mode from the MMoK policy as an illustration of the task.

\begin{widefigure}[t]
\centering
\includegraphics[width=\linewidth]{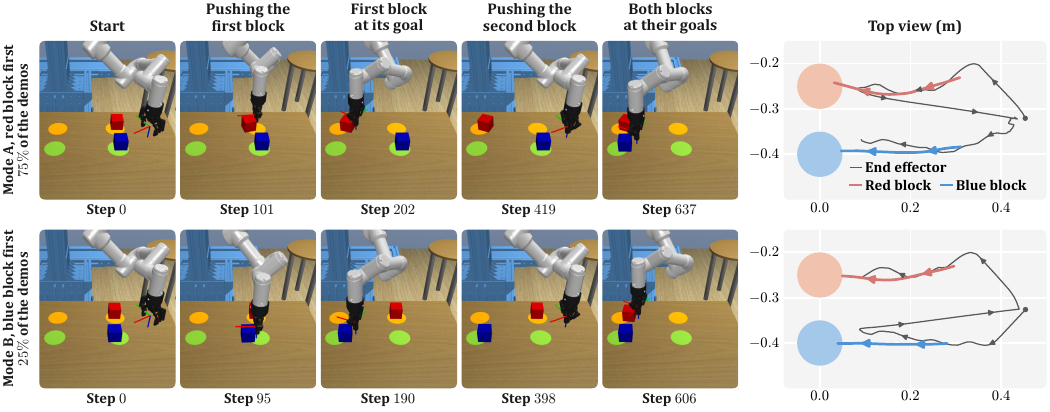}
\caption{Two successful rollouts of the MMoK policy on UR3 BlockPush ($\epsilon = 0.03$, $K = 20$, seed~$0$, epoch $150$), one for each mode.
In the top views, the shaded circles are the goal regions, and red and blue mark the two blocks.}
\label{fig:blockpush_modes}
\end{widefigure}

\subsection{Ambiguous medical image segmentation}
\label{app:medseg_examples}

\noindent\textbf{Experimental details}\par
\noindent\textit{LIDC-IDRI data.}
We use the LIDC-IDRI dataset~\citep{armato2011lidc}, which contains $15\,096$ lung CT crops of $128 \times 128$ pixels, each with the masks of four raters.
The crops come from a public PyTorch reimplementation of the Probabilistic U-Net~\citep{kohl2018probabilistic}.\footnote{\raggedright\url{https://github.com/stefanknegt/Probabilistic-Unet-Pytorch}}
The crops are split by patient $60/20/20$ into $9\,294$ training, $3\,222$ validation, and $2\,580$ test crops.

\noindent\textit{QUBIQ data.}
We use the brain-growth task of QUBIQ 2021~\citep{li2405qubiq}, which contains $39$ MRI slices, each with the masks of seven raters.
Because the challenge does not release its test labels, we combine its training and validation sets and evaluate with five-fold cross-validation by case, without a separate validation split.
Each image and its rater masks are zero-padded to a square and resized to $128 \times 128$ pixels.

\noindent\textit{Loss, training, and checkpoints.}
Like the $K$ candidates, the extra sample $\tilde y$ of MMoK is kept as a probability map.
Also, the mirror term applies the same soft Dice and binary cross-entropy to this probability map as the target, with no extra weight.
On LIDC-IDRI, validation GED$^2$ is measured every $1$k steps with $16$ samples per crop on a fixed random subset of $512$ validation crops.
For each method and $K$, the step $b$ in $2$k to $29$k with the lowest validation GED$^2$ averaged over the three seeds is selected, and the test score averages the EMA checkpoints at $b - 1$k, $b$, and $b + 1$k.
On QUBIQ, we have no validation split, so the test score averages the EMA checkpoints at $18$k, $19$k, and $20$k steps, which are fixed in advance.

\noindent\textit{Metrics.}
The model outputs per-pixel probabilities, and each sample is binarized at $0.5$.
GED$^2$ uses the distance $d = 1 - \mathrm{IoU}$ between two masks, with $d = 0$ for two empty masks.
The diversity ratio is the mean $d$ between distinct samples divided by the mean $d$ between distinct raters, each averaged over images.
NSD compares the majority masks of the samples and of the raters on the $128 \times 128$ grid at a tolerance of $2$ pixels, and it is $1$ when both masks are empty and $0$ when only one is.
The QUBIQ score averages Dice between thresholded sample and rater frequency maps over the thresholds $0.1, 0.2, \dots, 0.9$.

\section{Limitations and future work}
\label{app:limitations}
This paper introduces a few interesting open problems.
Our guarantees for MMoK concern stationary points.
\cref{thm:mirror} shows that the data distribution is a stationary point of MMoK training, and \cref{prop:mirror_atoms} shows that it is the only one in the simplified model of training on equidistant point modes.
We do not prove that training converges to it, which would require an analysis of the stochastic training dynamics.
Empirically, we evaluate MMoK on five scenarios, but not on large-scale generation of high-dimensional data such as natural images.
Future work could study the convergence of MMoK training and apply MMoK to larger-scale and higher-dimensional tasks.

\end{document}